\documentclass[10pt, final, journal, letterpaper, twocolumn]{IEEEtran}
\ifCLASSINFOpdf
\else
\fi

\usepackage[utf8]{inputenc}
\usepackage{amssymb}
\usepackage{hyperref}
\usepackage{bbm}
\usepackage{amsfonts}
\usepackage[dvips]{graphicx}
\usepackage{times}
\usepackage[]{cite}
\usepackage{amsmath}
\usepackage{array}
\usepackage{amssymb}

\usepackage{stfloats}
\usepackage{slashbox}
\usepackage{graphicx}
\usepackage{footnote}
\usepackage{booktabs}
\usepackage{array}
\usepackage{mathtools}
\usepackage{dsfont}
\usepackage{extarrows}
\usepackage{ntheorem}
\usepackage{subeqnarray}
\usepackage{cases}
\usepackage{threeparttable}
\usepackage{color}
\usepackage{nomencl}
\makenomenclature
\usepackage[numbers,sort&compress]{natbib}
\IEEEoverridecommandlockouts
\newtheorem{theorem}{Theorem}
\newtheorem{remark}{Remark}

\newtheorem{lemma}{Lemma}
\newtheorem{proof}{Proof}
\newtheorem{corollary}{Corollary}
\usepackage{flafter}
\usepackage{subfigure}
\usepackage{float}
\usepackage{bbold}
\usepackage{supertabular}
\usepackage{caption}
\usepackage{url}
\usepackage{xcolor}
\usepackage[ruled,vlined,lined,ruled,linesnumbered]{algorithm2e}
\newcommand{\bm}[1]{\mbox{\boldmath{$#1$}}}

\usepackage{subfigure}
\usepackage{stfloats}
\usepackage{enumitem}
\usepackage{etoolbox}
\renewcommand\nomgroup[1]{%
  \item[\bfseries
  \ifstrequal{#1}{P}{Physics Constants}{%
  \ifstrequal{#1}{N}{Number Sets}{%
  \ifstrequal{#1}{O}{Other Symbols}{}}}%
]}
\allowdisplaybreaks
\begin{document}
\title{\huge{Complex-Value Spatio-temporal Graph Convolutional
Neural Networks and its  Applications
to Electric Power Systems AI}}
\IEEEaftertitletext{\vspace{-2.2\baselineskip}}
\IEEEoverridecommandlockouts
\author{\IEEEauthorblockN{Tong~Wu,~\IEEEmembership{Member,~IEEE}}, 
\IEEEauthorblockN{Anna~Scaglione,~\IEEEmembership{Fellow,~IEEE}},
\IEEEauthorblockN{Daniel Arnold,~\IEEEmembership{Member,~IEEE}},\\
 \thanks{Tong Wu,  and Anna Scaglione are with the department of Electrical and Computer Engineering at Cornell Tech Campus,  Cornell University, NY, USA (e-mail: \{tw385, as337\}@cornell.edu). Daniel Arnold is with Lawrence Berkeley National Laboratory (e-mail: dbarnold@lbl.gov). }
%
}

\maketitle
\newcommand{\norm}[1]{\left\lVert#1\right\rVert}
\newcommand*\abs[1]{\lvert#1\rvert}

\begin{abstract}
The effective representation, precessing, analysis, and visualization of large-scale structured data over graphs are gaining a lot of attention. So far most of the literature considered exclusively real-valued signals. However, signals are often sparse in the Fourier domain, and more informative and compact representations for them can be obtained using the complex envelope of their spectral components, as opposed to the original real-valued signals.  
Motivated by this fact, in this work we generalize graph convolutional neural networks (GCN) to the complex domain, deriving the theory that allows to incorporate a complex-valued graph shift operators (GSO) in the definition of graph filters (GF) and process complex-valued graph signals (GS). The theory developed is generalized to handle spatio-temporal complex network processes.  We  prove that complex-valued GCNs can be stable with respect to perturbations of the underlying graph support, by bounding of the error propagation through multiple NN layers.  Then we apply complex GCN to power grid state forecasting, power grid cyber-attack detection and localization and demonstrate their superior performance relative to several benchmarks.
\end{abstract}

\section{Introduction}
 In machine learning (ML) applications in which signals have a sparse spectrum, the best representation for signals is through their complex envelopes. This explains the popularity of complex-valued neural networks  (Cplx-NN), introduced in the seminal paper \cite{leung1991complex}, in a number of different domains, such as physical layer communications, biological signals, array processing etc.
(see \cite{trabelsi2018deep} for a review of the theory behind  Cplx-NN and \cite{bassey2021survey} for a survey of its main applications). 
Motivated primarily by the application of  Artificial Intelligence (AI) in electric power systems, 
the overarching goal of this paper is to extend the benefits of Cplx-NN to the analysis of complex graph signals, first introducing Complex Graph Convolutional Neural Networks (Cplx-GCNs), and then investigating their potential benefits in processing electric power systems  measurements.
In fact, in power systems the AC voltage at each node (called bus) is concentrated around 60 or 50 Hz and the vector of complex envelopes of the voltage signals (called phasors) represents the state of the electric power network.  
The abundance of high-quality estimates of the voltage and current phasors acquired using phasor measurement units (PMUs) has already spurred interest in complex graph signal processing (GSP) as a framework to process them and interpret their properties \cite{ramakrishna2021gridgraph}.
Complex GSP is the most natural framework to analyze the power system state because it has a physical interpretation rooted in Ohm's law  \cite{ramakrishna2021gridgraph}.

GSP is a vibrant branch of signal processing research whose aim is to generalize digital signal process (DSP) notions, and Fourier analysis in particular, for data supported on graphs \cite{sandryhaila2013discrete}. 
A graph signal (GS) is a vector indexed by the node set of a weighted graph, representing both the data (node attributes) and the underlying structure (edge attributes). The cornerstone of GSP is the definition of Graph Shift Operator (GSO)\footnote{The name comes from the fact that originally the GSO was a generalization of the $z$ variable, corresponding to a time shift in the $z$ transform, although the definition often selects the Laplacian of the network graph, which is a graph signal differential operator.}. The vast majority of GSP-based algorithms uses real-valued GSOs and considers real-valued graph signals (GS) (see e.g. the surveys \cite{ortega2018graph, dong2020graph}). Having selected the GSO, one can define graph-filters; the most popular graph filter model is the Chebyshev filter \cite{sandryhaila2013discrete,defferrard2016convolutional, bronstein2017geometric}.
The development of complex valued GSP has received far less attention. In addition to power systems \cite{ramakrishna2020user, ramakrishna2021gridgraph}, complex GSP algorithms have been found applications, for example, in wireless communication networks \cite{adali2011complex} and sensor networks \cite{jablonski2017graph}. 
%
In power systems, complex GSP is the most natural framework \cite{ramakrishna2021gridgraph} since using the complex system matrix as the Graph Shift Operator (GSO) has a physical interpretation rooted in Ohm's law. The caveat is that the  GSO, which is the admittance matrix, is only symmetric, not conjugate symmetric, which, as we later discuss in Section \ref{sec:GGSP}, requires some special care.  



GSP algorithms that rely only on linear models have limited representation capability.
Interestingly, the first instance of Graph Neural Networks (GNN) architecture appeared well ahead of the development of GSP \cite{sperduti1997supervised}. The early models of GNN can be interpreted as a special case of the more general design introduced in \cite{bruna2014spectral}, where the authors extend the Convolutional NN (CNN) model using graph filters. 
To process time-series of graph signals, whose samples are not independent and identically distributed (i.i.d.),
the most effective architectures are  Spatio-Temporal versions of this idea, such as (STGCN) (see e.g. \cite{yan2018spatial,yu2018spatio} which are early works on the subject) and Graph Recursive NN (GRN) (first proposed in \cite{seo2018structured}). In a nutshell, their design includes feed-forward and feedback graph-temporal filters in each layer.
A thorough analysis of the stability of these designs is in \cite{gama2020stability}, which inspired the stability analysis in this paper.
Real-valued GCNs have showcased strong generalization capability in high-dimensional state spaces, learning complicated tasks with lower prior knowledge \cite{zhou2020graph}. 


To the best of our knowledge, thus far, GCNs (and its variants) have been studied and applied only in the real domain  (see e.g. \cite{zhou2020graph} for a review).
The construction of complex GCN we study in this paper follows exactly the same logic of cascading layers of complex graph-temporal filters with nonlinear activation functions for complex data. 
Spatio-temporal graph convolutional neural networks (Cplx-STGCN) are applicable not only to power systems, but to any networked system where nodal signals and their interactions can be modeled effectively as a vector of envelopes for its spectral components.
Prior to summarizing our contributions, next we  provide a brief review of the literature on real-valued GCN for power systems applications, including the ones that we consider in our experiments to test the Cplx-STGCN performance.  

\subsection{Related Works} 
Several papers have already applied real-valued GCN to power systems' data analysis and management \cite{liao2021review}. Applications include, for example, fault localization \cite{chen2019fault}, power  system  state  estimation \cite{zamzam2020physics}, anomaly detection \cite{cai2021structural, ma2021comprehensive}, detection and localization of stealth false data injection (FDI) attacks, synthetic feeder generation \cite{liang2020feedergan}, to name a few.  


The two applications we choose to test numerically Cplx-GCN architectures are that of detection and localization of FDI attacks and power systems state estimation and forecasting (PSSE and PSSF). We note that PSSF has so far been pursued  via  single-hidden-layers NNs \cite{do2009forecasting1, do2009forecasting2}, and further investigated by the Recurrent neural networks in \cite{zhang2019power} and Graph Recurrent neural networks \cite{hossain2021state}. The state-of-art neural network algorithms for FDI attack detection have been pursued by the Chebyshev GCN\cite{boyaci2021joint}, CNN \cite{wang2020locational} and RNN \cite{wang2021kfrnn}.

All works on real-GCN for power systems have in common the following limitations: 1) they ignore the correlation among real and imaginary parts of power systems signals and use real GSO; 2) they do not consider temporal correlation of voltage phasors samples.

\subsection{Contributions}
The aim of this paper is to establish the framework of complex-valued STGCN and elucidate how they can be applied to power grid signals inference problems. Our main contributions are as follows:
\begin{itemize}
	\item We combine the ideas in \cite{leung1991complex} and \cite{bruna2014spectral} and generalize the training of graph convolutional neural networks (GCN) so that they can operate complex domain, with complex-valued graph shift operators (GSO) and complex-valued graph signals.  
	\item We provide analytical bounds for the impact of perturbations in the GSO, and derive bounds for how the error propagates through the multi-layer GNN structure. 
	\item We further extend GCN to process streaming data through Cplx-STGCN architectures. 
	\item We show how to apply correctly this framework to power systems. This entails choosing as input the voltage phasors signals, the admittance matrix as our GSO, and the Graph Fourier basis suggested in \cite{ramakrishna2021gridgraph}. 
	\item We show that our method outperforms the prior art in  detecting and localizing FDI attacks as well as in PSSE and PSSF mean squared error (MSE) performance. We also empirically evaluate the sensitivity of the architecture to model changes.
	
	\end{itemize}

The rest of the paper is organized as follows. In Section II, we briefly review the key notions of GSP, setting the stage in Section III where we derive the physics inspired GSO and introduce our graph neural networks architectures whose   sensitivity is analyzed in Section IV. 
In Section V, we describe two applications of the proposed GCN and GRN frameworks that are tested numerically in Section VI. Finally, we conclude the paper in Section VII.

\section{Preliminaries on Power Systems}
The electric grid network has an associated undirected weighted graph $\mathcal{G}(\mathcal{V}, \mathcal{E})$ where nodes are \textit{buses} and its edges are its \textit{transmission lines}. The edge weights $\forall (i,j)\in {\cal E}$ are branch admittances $y_{ij} \in \mathbb{C}$ and each node has a shunt admittance $y_{ii}^{sh}, i\in {\cal V}$. With these parameters one can define the \textit{system admittance matrix} $\bm{Y} \in \mathbb{C}^{\abs{\mathcal{V}} \times \abs{\mathcal{V}} }$ that is defined as
\begin{equation}
[\boldsymbol{Y}]_{i, j}=\left\{\begin{array}{l}
y^{sh}_{ii} + \sum_{k \in \mathcal{N}_{i}} y_{i, k}, i=j \\
-y_{i, j}, i \neq j
\end{array}\right.
\end{equation}
Kichhoff's and Ohm's laws relate the current and voltage phasors for the entire network as follows:
\begin{eqnarray}
 \bm{i} = \bm{Y} \bm{v}, ~  v_{n} = \abs{v_{n}}~e^{\mathfrak{j} \varphi^v_{n}},~ 
     i_{n} =  \abs{i_{n}}~e^{\mathfrak{j} \varphi^c_{n}},~ \forall {n} \in \mathcal{V}, 
 \label{eq:node_injection}
\end{eqnarray}
where $\bm{v}$ and $\abs{\bm{v}}$ are the vectors of bus voltage phasors and magnitudes, respectively, with $\bm{v} \in \mathbb{C}^{ {  \abs{\mathcal{V}}}\times 1}$ and $\abs{\bm{v}} \in \mathbb{R}_+^{^{ {  \abs{\mathcal{V}}}\times 1}}$, $\mathfrak{j} = \sqrt{-1}$ is the  imaginary unit and $\bm{i} \in \mathbb{C}^{^{ {  \abs{\mathcal{V}}}\times 1}}  ~\textnormal{and}~\abs{\bm{i}}\in \mathbb{R}_+^{^{ { \abs{\mathcal{V}}}\times 1}}$ denote the vectors of net bus current phasors and magnitudes. Ohm's law allows us to view voltage as the output {\it low-pass} filter by ${\bm v} ={\bm Y}^{-1} \bm i$.
Let $\bm{s} = \bm{p} + \mathfrak{j} \bm{q}$ be the vector of net apparent power at buses ($\bm{s} = [{s}_1, \cdots, {s}_{\abs{\mathcal{V}}}]^\top$),  with the $n^{th}$ entry  ${s}_n = {p}_n + \mathfrak{j} {q}_n$, where ${p}_n$ and ${q}_n$ are the active and reactive power injection at bus $n$, respectively.  

The vector of net apparent power injections is:
\begin{eqnarray}
\bm{s} = \bm{v} \odot \bm{i}^* = \bm{v}  \odot (\bm{Y} \bm{v})^*, \label{sinj}
\end{eqnarray}
where $\odot$ is the Hadamard product and $\bm{i}^*$ is the conjugate of a complex vector $\bm{i}$. 
$\bm{s} = \bm{p} + \mathfrak{j} \bm{q}$ and the $n^{th}$ entries of the real and imaginary parts ${p}_n$ and ${q}_n$ are the active and reactive power injection at bus $n$, respectively.  

The appropriate grid graph shift operator (GSO) $\mathbf{S}$ is the system admittance matrix, $\mathbf{S} = \bm{Y}$. Note that unlike the graph Laplacian, this GSO $\mathbf{S} = \bm{Y}$ is invertible (albeit ill-conditioned), thanks to the diagonal component of the shunt admittances\footnote{The magnitudes of the shut admittaces $y^{sh}_{ii}$ are very small relative to the line admittances $y_{ij}$ and are often neglected}. 

\section{Complex-Valued Spatio-Temporal Graph Convolution Neural Network}
\subsection{Grid-Graph Signal Processing}\label{sec:GGSP}
In our work the graph signal $\bm{x}  \in \mathbb{R}^{\abs{\mathcal{V}}}$ is a vector of voltage phasors at each bus, an $\left[\bm{x}\right]_i, \forall i \in \mathcal{V}$ is the $i$-th entry of this state vector.
The set $ \mathcal{N}_i$ denotes the subset of nodes connected to node $i$.  A graph shift operator (GSO) is a matrix $\mathbf{S} \in \mathbb{R}^{\abs{\mathcal{V}} \times \abs{\mathcal{V}}}$ that linearly combines graph signal neighbors' values. Almost all operations including filtering, transformation and prediction are directly related to the GSO. 
In this work, we focus on complex symmetric GSOs, i.e. such that $\mathbf{S} = \mathbf{S}^\top$ that are appropriate for our power grid application where $\mathbf{S} = \bm{Y}$. 
A graph filter is a linear matrix operator $\mathcal{H}(\mathbf{S})$ that is a function of the GSO and operates on graph signals as follows  
\begin{align}\label{gs}
    \bm{w} = \mathcal{H}(\mathbf{S}) \bm{x}.
\end{align}
What defines the dependency of $\mathcal{H}(\mathbf{S})$ on the GSO is that  $\mathcal{H}(\mathbf{S})$ must be shift-invariant (like a linear
time invariant filter in the time domain), i.e. matrix operators such that $\mathbf{S}\mathcal{H}(\mathbf{S})\equiv \mathcal{H}(\mathbf{S})\mathbf{S}$. This property is satisfied if and only if $\mathcal{H}(\mathbf{S})$ is a matrix polynomial:
\begin{equation}
\begin{split}\label{lsi}
 \mathcal{H}(\mathbf{S}) = \sum_{k=0}^{K-1} h_k  \mathbf{S}^k.
\end{split}
\end{equation}
where the graph filter order $K$  can be infinite.
Let the eigenvalue decomposition be $\mathbf{S} = \mathbf{U}\bm{\Lambda} \mathbf{U}^\top$ where $\bm{\Lambda}$ is a diagonal matrix with eigenvalues $\abs{\lambda_1}\leq \abs{\lambda_2}\leq \dots\leq\abs{\lambda_{\abs{\mathcal{V}}}}$ and $\mathbf{U}$ be the eigenvector matrix that is unitary since the GSO $\bf S$ is symmetric.\footnote{Note that   the graph frequencies of complex GSO are denoted by  $\abs{\lambda_n}$, which tend to be unique \cite{horn2012matrix, ramakrishna2021gridgraph}.}  The Graph Fourier Transform (GFT) basis is $\mathbf{U}$, the GFT of a graph signal is $\tilde{\bm x}=\mathbf{U}^\top\bm x$ and the eigenvalues $\lambda_{\ell}, \ell=1,\ldots, \abs{\mathcal{V}}$ are the \textit{graph frequencies}. 
From \eqref{lsi} it follows that:
\begin{equation}
\begin{split}
\mathcal{H}(\mathbf{S}) = \mathbf{U} \left(\sum_{k=0}^{K-1} h_{k} \bm{\Lambda}^k \right) \mathbf{U}^{\top}.
\end{split}
\end{equation}
The matrix $\sum_{k=0}^{K-1} h_{k} \bm{\Lambda}^k$  is a diagonal, with $i^{th}$ entry $\tilde{h}(\lambda_i) \triangleq \sum_{k=0}^{K-1} h_k \lambda^k_i $.  Hence, $\tilde{\bm h}=[\tilde{h}(\lambda_1),\ldots,\tilde{h}(\lambda_{|\cal V|})]$ is the transfer function for graph filters, and in the GFT domain, and the graph filter output corresponds to an element by element multiplication of the graph filter input, i.e.:
\begin{equation}
\begin{split}
\bm{w} = \mathcal{H}(\mathbf{S})\bm{x}  \iff \tilde{\bm{w}} = \tilde{\bm{h}} \odot \tilde{\bm{x}},
\end{split}
\end{equation}
where and $\odot$ represents the element by element (Hadamard) vector product.  
To process time series of graph signals $\{\bm{x}_t\}_{t\ge 0}$ with samples that are not i.i.d., graph temporal filters models are more appropriate:
\begin{equation}
\bm{w}_t  = \sum_{\tau = 0}^t \mathcal{H}_{t - \tau}( \mathbf{S}  ) \bm{x}_\tau~~~~\mathcal{H}_t(\mathbf{S}) = \sum_{k=0}^{K-1} h_{k,t} \mathbf{S}^k,
\end{equation}
and for their analysis we can harness DSP tools, defining a combined GFT and $z-$transform:
\begin{equation}
\begin{split}
\mathbf{X}(z) = \sum_{t=0}^{K_t-1} \bm{x}_t z^{-t}, ~~~\tilde{\mathbf{X}}(z) = \mathbf{U}^\top\mathbf{X}(z),
\end{split}
\end{equation}
where $K_t$ is the length of the graph signal time series. 
In particular, considering a filter of order $K_t$, we use $\mathbf{S} \otimes z$ ($\otimes$ is tensor product) as the graph temporal GSO:
\begin{eqnarray}
\mathcal{H}(\mathbf{S} \otimes z) = \sum_{k=0}^{K-1} H_k(z) \mathbf{S}^k,~~
 H_k(z)=\sum_{t=0}^{K_t-1} h_{k,t}z^{-t}
\end{eqnarray}
i.e. $H_k(z) $ is the $z-$transform of the filter coefficients $h_{k, t}$.
 In the $z$-domain, the input-output relationship is expressed as:
\begin{equation}
\mathbf{W}(z) = \mathcal{H}(\mathbf{S} \otimes z) \mathbf{X}(z). 
\end{equation}
The graph-temporal transfer function and input-output relationship in the joint GFT-$z$-domain are: 
\begin{equation}
\begin{split}\label{stkernal}
\mathbb{H}(\bm \Lambda, z) = \sum_{t=0}^{K_t-1} \sum_{k=0}^{K-1} h_{k, t} \bm \Lambda^k z^{-t}, ~~~ \tilde{\mathbf{W}}(z) =\mathbb{H}(\bm \Lambda, z) \tilde{\mathbf{X}}(z),
\end{split}
\end{equation}
where $\mathbb{H}(\bm \Lambda, z) $ is a diagonal matrix and  $\tilde{\mathbf{X}}(z)=\mathbf{U}^\top \mathbf{X}(z)$,
which is again an element by element multiplication since $\mathbb{H}(\bm \Lambda, z)$ is a diagonal matrix. 
In a graph-convolutional neural networks (GCN), the coefficients $h_{k, t}$ are the trainable parameter \cite{defferrard2016convolutional}.  The subtle differences between complex and real GSP have been discussed in \cite{ramakrishna2019modeling}. 

\subsection{Complex-valued Graph Convolution Neural Network}
The graph neural network perceptron based on \eqref{gs} is:
\begin{equation}
\bar{\bm{w}}=\sigma[\bm{w}] = \sigma\left[\sum_{k=0}^{K-1} h_{ k} \mathbf{S}^{k} \bm{x}\right]
\end{equation}
where $\bm{x}\in \mathbb{C}^{\abs{\mathcal{V}}}$, $h_k\in \mathbb{C}$, $\mathbf{S}^k\in \mathbb{C}^{\abs{\mathcal{V}}\times \abs{\mathcal{V}}}$ and $\bm{w}\in \mathbb{C}^{\abs{\mathcal{V}}}$ are the complex values. Since, $\sigma(\cdot)$ takes as input complex values, there is significant flexibility in defining this operator in  the complex plane. In the following, we refer to  Complex ReLU (namely CReLU) as the simple complex activation that applies separate ReLUs on both of the real and the imaginary part of a neuron, i.e: 
\begin{equation}
\operatorname{CReLU}(\bm{w})=\operatorname{ReLU}(\Re(\bm{w}))+\mathfrak{j} \operatorname{ReLU}(\Im(\bm{w})).
\end{equation}
This is a popular choice because the CReLU satisfies the Cauchy-Riemann equations if both the real and imaginary parts are either strictly positive or strictly negative \cite{trabelsi2018deep}. Empirically, we found this choice to be preferable to other options proposed in the literature. 

Spatio-Temporal GCN are a special case of {\it multiple features GCN}. Specifically, let $\mathbf{X} = [\bm{x}^1, \cdots, \bm{x}^F]$ and let us refer to the multiple channel outputs as $\mathbf{W} = [\bm{w}^1, \cdots, \bm{w}^G]$, where $F$ is the number of input features and $G$ is the number of output channels. A layer of  multiple features GCN operates as follows:
\begin{equation}\label{eq:MF-GCN}
\bar{\mathbf{W}} = \sigma[\mathbf{W}] =\sigma\left[\sum_{k=0}^{K-1} \mathbf{S}^{k} \times \mathbf{X} \times \mathbf{H}_{k}\right] = \operatorname{CReLU}(\mathbf{H}*_{\mathcal{G}} \mathbf{X}),
\end{equation}
where these matrices include $ G \times F$ coefficient matrix $\mathbf{H}_{k}$ with entries $\left[\mathbf{H}_{k}\right]_{{fg}}=h_{k}^{f g}$, and  $\mathbf{H}*_{\mathcal{G}}$ defines   the notion of graph convolution operator based on the concept of spectral graph convolution.

\subsubsection{Discussion about Cplx-STGCN vs Real-STGCN}
 
Note that using real-GCN in lieu of complex GCN
reduces significantly the number of trainable  parameters. Specifically,  in terms of Cplx-GCN,  one way of mixing and separating the real and imaginary variables is
 \begin{equation}
\begin{bmatrix}
\Re(\bm{w})\\
\Im(\bm{w})
\end{bmatrix} =  \sum_{k=0}^{K-1} h_{ k} \begin{bmatrix}
\Re(\mathbf{S})&-\Im(\mathbf{S})\\
\Im(\mathbf{S})&\Re(\mathbf{S})
\end{bmatrix}^{k} \begin{bmatrix}
\Re(\bm{x})\\
\Im(\bm{x})
\end{bmatrix} 
\end{equation}
 using an $ h_{ k}$ which is a real scalar in the decoupled model. This removes the imaginary part of $ h_{ k}$, reducing the neural network function approximation capability. This is why, when such GCN methods are applied to voltage phasor signals, the resulting  trained models under-perform the complex ones in inference and control tasks.

\section{Analysis of the Cplx-GCN sensitivity}
Particularly for power systems, it is quite common to incur in sparse system changes, due to switching or changes of line impedance. It is, therefore, of interest to understand how sensitive is the response of the Cplx-GCN to changes in the parameters. For the case where the changes in the GSO are known it is to study how parameters  trained on a different GSO will respond. We refer to this as the {\it transfer learning} error.
Next we  provide insights on the impact of perturbations in the GSO on the end-to-end  Cplx-GCN mapping. We improve substantially the results of \cite{gama2020stability} which  are exclusively for real GNN, do not consider the end to end distortion, and also rely on restrictive assumptions about the structure of the perturbation that we could not justify in our practical setting.

In the following we denote by $\sigma_{\max}(\mathbf{A})$ its largest singular value of matrix $\mathbf{A}$. 
We can prove the following bound:
\begin{theorem}\label{transferbound}
Consider graph filter $\bm{h} = [h_0, \cdots, h_{K}]$ along with shift operator $\bf{S}$  having $\abs{\mathcal{V}}$  nodes. Let  $\mathbf{E} \in \mathbb{C}^{\abs{\mathcal{V}}\times \abs{\mathcal{V}}}$ denote the matrix perturbation with $\norm{\mathbf{E}} \le  {\epsilon}$, and $\hat{\bf{S}} = \bf{S} + \mathbf{E}$. Let us denote by  $\hat{\bm{h}} = [\hat{h}_0, \cdots, \hat{h}_{K}]$ 
the graph filter parameters obtained training the network using $\hat{\mathbf{S}}$ as GSO. Let $\hat{\mathcal{H}}(\hat{\mathbf{S}}) = \sum_{k= 0}^K \hat{h}_k  \hat{\bf{S}}^k $ and the cplx-GCN layer with the  the original filter coefficients obtained training with GSO $\mathbf{S}$, albeit using the perturbed GSO, be $ \mathcal{H}(\hat{\mathbf{S}}) = \sum_{k= 0}^K   h_k  \hat{\bf{S}}^k $. Let us also define:
\begin{align}
\gamma_1&\triangleq \max\Big(1, (\sigma_{\max}(\mathbf{S}) +\epsilon )^K  \Big) \label{eq:gamma1}
\end{align} 
The following bound holds:
\begin{equation}
\begin{aligned}
&	\sigma_{\max}(\hat{\mathcal{H}}(\hat{\mathbf{S}}) -  {\mathcal{H}}(\hat{\mathbf{S}}))   \le   \gamma_1
\|\hat{\bm{h}}- {\bm{h}}
 \|_{1} .
\label{eq:Thm1bound}
\end{aligned}
\end{equation}
\end{theorem}
\begin{proof}
The proof is in Appendix \ref{app:Thm1}.
\end{proof}

\begin{theorem}\label{th2}
	Let $\mathbf{E} \in \mathbb{C}^{\abs{\mathcal{V}}\times \abs{\mathcal{V}}}$ be the matrix perturbation with $ \norm{\mathbf{E}} \le \epsilon $, and $\hat{\bf{S}} = \bf{S} + \mathbf{E}$ and
\begin{equation}
\gamma_2\triangleq 
\max_{1\leq k\leq K}|h_k| (1+\sigma_{\max}(\mathbf{S}))^K.
\label{coeff2}
\end{equation} 
	Assume that the cplx-GCN layer used is  $ {\mathcal{H}}(\hat{\mathbf{S}}) = \sum_{k= 0}^K  {h}_k  \hat{\bf{S}}^k $ where the coefficients are the same as those obtained by training in the original cplx-GCN layer is defined as $ {\mathcal{H}}(\mathbf{S}) = \sum_{k= 0}^K  {h}_k   {\bf{S}}^k $. Then, the following  bound holds:
\begin{equation}
\begin{aligned}\label{theo2}
& \sigma_{\max}( {\mathcal{H}}(\hat{\mathbf{S}}) -  {\mathcal{H}}  ( {\mathbf{S}}))  \le    \gamma_2 \frac{{\epsilon}(1-{\epsilon}^K)}{1- {\epsilon}}.
\end{aligned}
\end{equation}
\end{theorem}
\begin{proof}
The proof is in Appendix \ref{app:Thm2}.
\end{proof}
This theorem clearly shows that for a small perturbation in the GSO one should get a similar response to the parameters, which suggests that for small GSO perturbations is reasonable to use the same parameters and transfer the learning done to the new case. 

Next we bound the difference between the two outputs of the retrained network with the perturbed GSO and the original network, in other words the bound on the norm of the output difference, when the GCN perturbation    consists of  the perturbations of both parameters $h_k$ and GSO.
\begin{corollary}[The Bound of Cplx-GCN Perturbation]
Let the retrained cplx-GCN layer be $\hat{\mathcal{H}}(\hat{\mathbf{S}}) = \sum_{k= 0}^K \hat{h}_k  \hat{\bf{S}}^k $ and the original one be $ \mathcal{H}(\mathbf{S}) = \sum_{k= 0}^K   h_k   {\bf{S}}^k $ with the new GSO $\hat{\bf{S}}$.
Then:
	\begin{equation}
	\begin{aligned}\label{cor1}
	\|\big(
	\hat{\mathcal{H}}(\hat{\mathbf{S}})&-\mathcal{H}(\mathbf{S}) \big)\bm x
	\|\le\\ &\left(
	\gamma_1
\|\hat{\bm{h}}- {\bm{h}} \|_1  +
\gamma_2 \frac{{\epsilon}(1-{\epsilon}^K)}{1- {\epsilon}}
	\right)\|\bm x\|
	\end{aligned}
\end{equation}
where the parameters in the right-hand side were defined in Theorem 1 and 2.
\end{corollary}
The proof is obvious because of the triangle inequality.
\begin{remark}
If the GSO $\mathbf{S}$ is normalized by $\sigma_{\max}(\mathbf{S})$, we can further bound the result in \eqref{eq:Thm1bound} as follows:
\begin{equation}
\begin{aligned}
\gamma_1
\|\hat{\bm{h}}- {\bm{h}} \|_1 & \le  (1+\epsilon)^K \|\hat{\bm{h}}- {\bm{h}} \|_1	\end{aligned}
\end{equation}
$\gamma_1$ amplifies the sensitivity exponentially $K$ while the effect of the parameters difference increases linearly with $K$. When $K \rightarrow \infty$, we have
\begin{equation}
\begin{aligned}
\lim_{K\rightarrow \infty}(1+\epsilon)^K \|\hat{\bm{h}}- {\bm{h}} \|_1  = e^{\epsilon K} \|\hat{\bm{h}}- {\bm{h}} \|_1
	\end{aligned}
\end{equation}
Moreover, if $\epsilon $ is approaching 0, i.e., $\epsilon \rightarrow \frac{1}{K}$,  the error can be further bounded by 
\begin{equation}
\begin{aligned}
\lim_{K\rightarrow \infty, \epsilon\rightarrow \frac{1}{K}}(1+\epsilon)^K \|\hat{\bm{h}}- {\bm{h}} \|_1  = e \|\hat{\bm{h}}- {\bm{h}} \|_1
	\end{aligned}
\end{equation}

\end{remark}
%
\begin{remark}
The interpretation of Theorem \ref{th2} is more straightforward. Its dependence on $\epsilon$ is clear, and it is also clear that it exponentially increases with $K$ with a rate $(1+\sigma_{\max}(\bm S))$. So, when using an incorrect GSO in the network one would want to make sure that $\epsilon \ll (1+\sigma_{\max}(\mathbf{S}))^{-K}$. Here,  $\sigma_{\max}(\mathbf{S})$ and $K$ are clearly negatively impacting the sensitivity. 
\end{remark}


\subsection{Spatiotemporal Graph Convolutional Neural Network}
\begin{figure}[!htp]
 \vspace{-0.4cm}
  \center
    \includegraphics[width=0.42\textwidth]{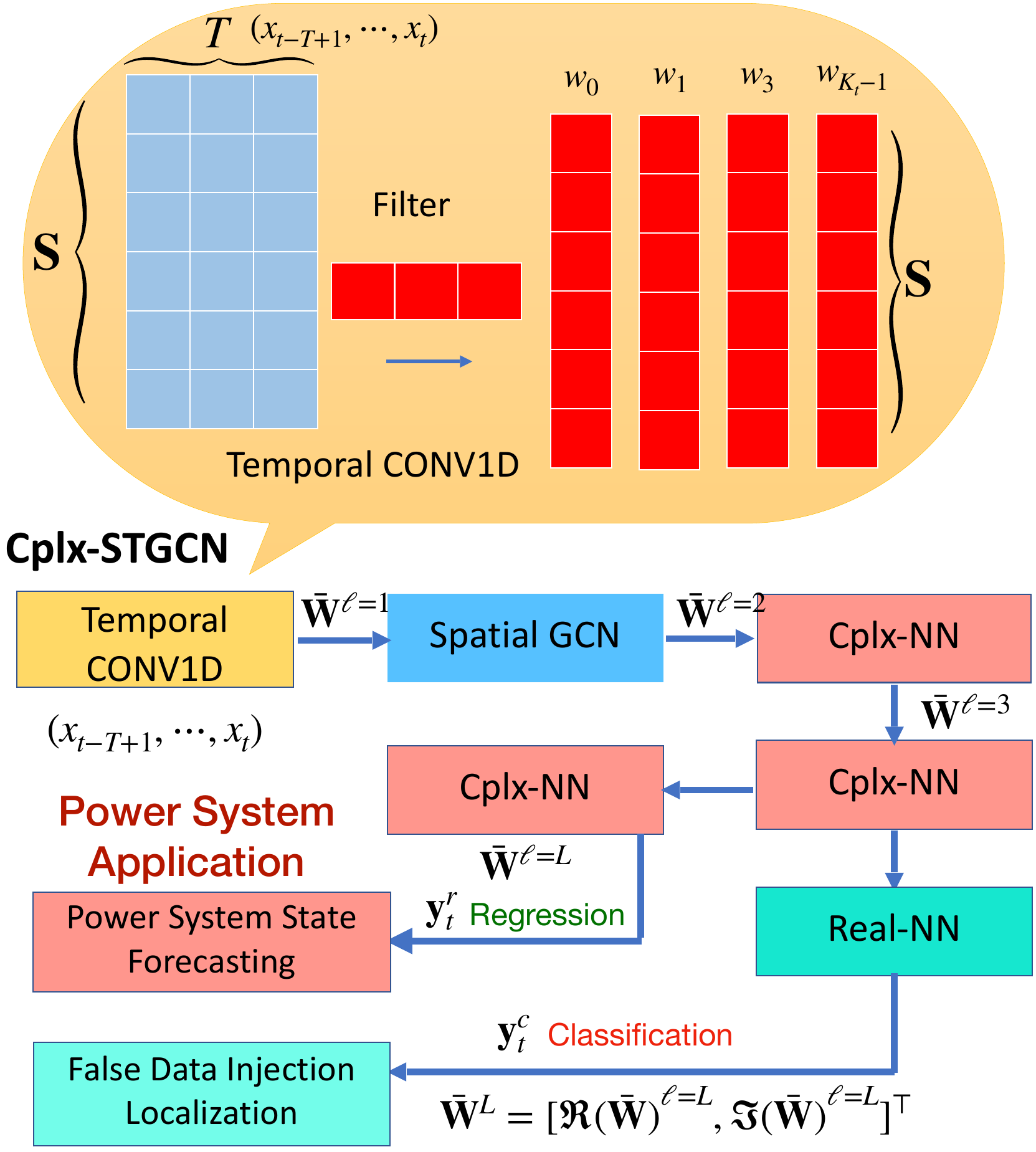}
  \caption{The Architecture of Cplx-STGCN.}\label{architecture_cgcn}
  \vspace{-0.4cm}
\end{figure}

Power systems are dynamic systems with time-varying voltage phasors. In order to fuse features
 from both spatial and temporal domains, we will consider the following two graph neural network architectures, namely Conv1D Graph convolutional neural networks.
Although RNN-based models become widespread in time-series analysis, RNN for power systems still suffers from time-consuming iterations, complex gate mechanisms, and slow response to dynamic changes. CNNs, on the other hand, allow for fast training, have a simpler structures, and no dependency constraints from previous steps. As shown in Fig. \ref{architecture_cgcn}, the temporal convolutional layer contains a 1-D CNN with a width-$T$ kernel with $K_t$ output channels. The convolutional kernel $\mathbf{\Gamma}\in \mathbb{C}^{T\times K_t}$ is designed to map the input $\mathbf{X}\in \mathbb{C}^{\abs{\mathcal{V}}\times T}$ into a output graph signal with $C_t$ channels $\bar{\mathbf{X}}\in \mathbb{C}^{\abs{\mathcal{V}}\times K_t}$. Therefore, we define the temporal convolution as,
\begin{equation}
 \bar{ \mathbf{X}} =  \mathbf{\Gamma}*_{\mathcal{T}} \mathbf{X},
\end{equation}
where each column of  $ [\bar{ \mathbf{X}}]_\tau$ is defined as $ \bar{ \bm{x}}_\tau, \tau = 0, 1, \cdots, K_t-1$.
After the temporal convolutional layer, we are ready to put $ \bar{ \mathbf{X}}$ into the spatial layer.
Based on \eqref{stkernal}, we can design the following transfer functions and neuron:
\begin{equation}
\begin{split}\label{stkernal2}
\mathbb{H}(\mathbf{S}, z) = \sum_{t=0}^{K_t-1} \sum_{k=0}^{K-1} h_{k, t} \mathbf{S}^k z^{-t},\\
\bar{\bm{w}}_t = \sigma[\bm{w}_t] =  \sigma\left[\sum_{k=0}^{K-1} \sum_{\tau=0}^{K_t-1} h_{k, \tau} \mathbf{S}^k \bm{x}_{t-\tau}\right].
\end{split}
\end{equation}
Accordingly, the graph signal $\bar{\bm{w}}_t$ from the spatial feature extraction layer (see Fig. \ref{architecture_cgcn}) is:
\begin{equation}
\begin{split}\label{graph_signal}
\bar{\bm w_t} =  \operatorname{CReLU} \left[\sum_{k=0}^{K-1} \sum_{\tau=0}^{K_t-1} h_{k, \tau} \mathbf{S}^k \bar{\bm{x}}_{t-\tau}\right],
\end{split}
\end{equation}
By combining the temporal and spatial convolutions at each layer,  the multiple output channels of the Cplx-STGCN layer ($\ell = 1$) are expressed as
\begin{equation}
\begin{split}\label{graph_signal}
\bar{\mathbf{W}}_{t,\ell=1}  = \operatorname{CReLU}(\mathbf{H}*_{\mathcal{G}} (\mathbf{\Gamma}*_{\mathcal{T}} \mathbf{X}_t)),
\end{split}
\end{equation}
where $\mathbf{H}$ and $\mathbf{\Gamma}$  are the trainable parameters. We  denote \eqref{graph_signal} by the feature extraction  layer.

The following hidden layers $\ell\in\{1, \cdots, L-1\}$    are the complex-valued fully connected neural network:
\begin{equation}
\begin{split}\label{cgcn_policy}
&\bar{\mathbf{W}}_{t,\ell+1}=  {\operatorname{CReLU}\left(\Theta_\ell^{cplx} * {\bar{\mathbf{W}}_{t,\ell}}\right)}, \quad L-1\ge\ell \ge 1.
\end{split}
\end{equation}
For the output layer $L$, we transform the complex tensor $\bar{\mathbf{W}}_{t,L}$ into a real tensor, and then  map it to the labels (or regression targets):
\begin{equation}\label{cgcn_policy}
\begin{aligned}
\text{regression:} ~\bm{y}^r = \text{tanh}\left(\Theta_{L}^{cplx} *
\bar{\mathbf{W}}_{t,L}
\right),\\
\text{classification:} ~  \bm{y}^c = \text{sigmoid}\left(\Theta_{L}^{re} *
\begin{bmatrix}
\Re(\bar{\mathbf{W}}_{t,L})\\
\Im(\bar{\mathbf{W}}_{t,L})
\end{bmatrix}
\right).
\end{aligned}
\end{equation}
where $\bm{y}^r$ and $\bm{y}^c$ denote the complex regression targets  and the real classification labels, respectively. Besides, $\Theta_{L}^{cplx}$ and $\Theta_{L}^{re}$ denote the complex and real trainable matrix.
Then, we define the multi-layer Cplx-STGCN learning function as: 
\begin{equation}
\begin{aligned}
&\text{regression:}~    \bm{y}^r_t = \Phi^r(\mathbf{X}_t, \mathbf{S}, \theta^r ),\\
&\text{classification:}~    \bm{y}^c_t = \Phi^c(\mathbf{X}_t, \mathbf{S}, \theta^c ),\label{predictlabel}
\end{aligned}
\end{equation}
where $\theta^r \triangleq \{(\Theta_{\ell}^{cplx}, \mathbf{H}, \mathbf{\Gamma})|\forall \ell = 1,\cdots, L\}$ and  $\theta^c \triangleq \{(\Theta_{\ell}^{cplx},\Theta_{L}^{real}, \mathbf{H}, \mathbf{\Gamma})|\forall \ell = 1,\cdots, L-1\}$ represent the trainable parameters and $\mathbf{X}_t = [\bm{x}_{t-T+1}, \cdots, \bm{x}_{t}]$. Here, we have omitted the bias term to unburden the notation, but they are present in the trainable model we use. 

%

In the following, we further investigate how the multilayer neural networks propagate the error due to the changes of parameter  and  GSO $\mathbf{S}$.
\begin{lemma}
	Assume a neural network consists of one Cplx-GCN layer and one Cplx-FNN layer with trainable parameters $\Theta^{cplx}$, denoted by $\bm{y} = \Phi(\bm{x}, \mathbf{S}, \theta) $. The retrained neural network has the new GSO $\hat{\mathbf{S}}$ and the new parameters $\hat{\theta}$, denoted by $\hat{\bm{y}} = \Phi(\bm{x}, \hat{\mathbf{S}}, \hat{\theta}) $. We define the perturbation of the Cplx-FNN layer is $\norm{\Theta^{cplx} - \hat{\Theta}^{cplx}}_2\le \delta_{\mathbf{w}}$. Then, the distance between ${\bm{y}}$ and $\hat{\bm{y}}$ is  bounded by:
 \begin{equation}
\begin{aligned}\label{2layerboundsre}
	\norm{{\bm{y}} - \hat{\bm{y}}} \le \Big(\overbrace{\delta_{\mathbf{w}} * \Psi_1 + \sigma_{\max}(\Theta^{cplx}) * \Psi_2}^{\triangleq {\Delta}_1}\Big) \norm{\bm{x}}
\end{aligned}
 \end{equation}
 where  $\Psi_1$  and $\Psi_2 $ are given by 
  \begin{equation}
\begin{aligned}\label{Psi1}
	& \Psi_1  =   \gamma_1\|  {\hat{\bm h} } \|_{1}\\
	& \Psi_2 =  \left[ 	\gamma_1\left\| \hat{\bm{h}}- {\bm{h}}
\right \|_{1}  + \gamma_2 \frac{{\epsilon}(1-{\epsilon}^K)}{1- {\epsilon}}\right] 
\end{aligned}
 \end{equation}
\end{lemma} 
\begin{proof}
The proof is in Appendix \ref{app:Thm3}.
\end{proof}
\begin{corollary}\label{cor2}
We generalize the bound into the   multilayer neural networks, including one Cplx-GCN feature extraction layer and $L$ cplx-FNNs as
	\begin{equation}
	\begin{aligned}
&\norm{{\bm{y}} - \hat{\bm{y}}} \le \Delta_L, ~~\Delta_L = \sigma_{\max}(  \Theta^{cplx}_L )\Delta_{L-1} \\
&  +   \delta_{\mathbf{w}_L} \prod_{\ell=1}^{L-1}\sigma_{\max}(  \hat{\Theta}^{cplx}_{\ell} )  \Psi_1.
		 \end{aligned}
\end{equation}
where $\Delta_1$ is defined in \eqref{2layerboundsre}.
\end{corollary}
The proof is obvious according to the norm triangle inequality so that we omit the proof. 
From Corollary \ref{cor2}, we can observe that if the retrained neural networks have small changes to make $  \delta_{\mathbf{w}_L} \approx 0$, we could find the dominant part, i.e., $\Delta_L \approx \sigma_{\max}(  \Theta^{cplx}_L )\Delta_{L-1}$. Therefore, the error of the GCN perturbation is propagated by the largest singular values of original cplx-FNN weight matrices.

\section{Applications of Cplx-STGCN}
\subsection{Power System State Estimation and Forecasting}
Measurements in power systems are relatively sparse. 
In this subsection, we propose a Power Systems State Estimation (PSSE) algorithm that can use limited measurements to estimate the current and future  state at all buses.
Let $\bm x_{\mathcal{A}}$ (the time index $t$ is ignored for simplicity) be the vector of measurements in the subset of buses $\mathcal{A} \subset \mathcal{V}$ that have sensors. 
Let the GFT basis corresponding  to the dominant $k$ graph frequencies in the voltage phasor GFT spectrum  $\mathbf{U}_{\mathcal{K}}$. Because of Ohm's law, this set corresponds to the lowest graph frequencies  \cite{ramakrishna2021gridgraph} and the best subset of measurement buses ${\cal A}$ is one-to-one with the subset of rows of $\mathbf{U}_{\mathcal{K}}$ with minimum correlation.
Let $ {\mathcal{F}}_\mathcal{A}$ be the so called {\it vertex limiting operator} i.e. the matrix such that $ {\mathcal{F}}_\mathcal{A} =  {\mathcal{Q}}_\mathcal{A}  {\mathcal{Q}}^\top_\mathcal{A}$, where $ {\mathcal{Q}}_\mathcal{A}$ has columns that are the coordinate vectors pointing to each vertex/node in ${\cal A}$, so that $\bm x_{\cal A}={\mathcal{Q}}_\mathcal{A}\bm x$.  Mathematically, the optimal placement can be sought by maximizing the smallest  singular value, $\max_{{\mathcal{F}}_{\mathcal{A}}}\varpi_{\min} ({\mathcal{F}}_{\mathcal{A}} \mathbf{U}_{\mathcal{K}})$, of the matrix ${\mathcal{F}}_{\mathcal{A}} \mathbf{U}_{\mathcal{K}}$. Such choice amounts to the selection of rows of $\mathbf{U}_{\mathcal{K}}$ that are as uncorrelated as possible, because the resulting matrix ${\mathcal{F}}_{\mathcal{A}} \mathbf{U}_{\mathcal{K}}$ has the highest condition number \cite{anis2016efficient}.

After choosing, using the aforementioned method, the best location for measuring the voltage phasors\footnote{These measurements can be collected by Phasor Measurement Units.} $\mathcal{A}$ the available measurements for the vector observation $ \bm{z}_t $.  
%
With $\mathcal{A}$ denoting the set of available measurement, we use $\mathcal{U}$ to denote the set of  unavailable ones. Therefore, \eqref{eq:node_injection} can be written as:
\begin{equation}
\underbrace{\left[\begin{array}{l}
\hat{\boldsymbol{i}}_{\mathcal{A}} \\
\hat{\boldsymbol{v}}_{\mathcal{A}}
\end{array}\right]}_{\boldsymbol{z}_{t}}=\underbrace{\left[\begin{array}{cc}
\boldsymbol{Y}_{\mathcal{A A}} & \boldsymbol{Y}_{\mathcal{A} \mathcal{U}} \\
\mathbb{I}_{|\mathcal{A}|} & \mathbf{0}
\end{array}\right]}_{\boldsymbol{H}} \underbrace{\left[\begin{array}{l}
\boldsymbol{v}_{\mathcal{A}} \\
\boldsymbol{v}_{\mathcal{U}}
\end{array}\right]}_{\boldsymbol{x}_t}+\boldsymbol{\varepsilon}_{t},
\end{equation}
where $\boldsymbol{\varepsilon}_{t}$ is a vector of measurement noise.  The voltage phasor forecasting is a typical time-series prediction problem, i.e. predicting the most likely voltage phasor measurements in the next $H$ time steps given the  previous  $T$ sub-sampled measurement $[\bm{x}_t]_{\mathcal{A}}$ as
\begin{equation}
\begin{aligned}
{\bm x}^*_{t+H}=  \underset{ \bm{x}_{t+H}}{\arg \max } \log P\left( \bm{x}_{t+H} \mid [\bm{x}_{t-T+1}]_{\mathcal{A}}, \ldots, [\bm{x}_{t}]_{\mathcal{A}}\right),
\end{aligned}
\end{equation} 	
where $[\bm{x}_{t}]_{\mathcal{A}} \in\mathbb{C}^{\abs{\mathcal{A}}}$ is an observation vector of $\abs{\mathcal{A}}$  measurements at time step $t$, each element of which records historical observation for a bus.

\subsubsection{Methodology}
 The first step of the algorithm is to recover the voltage phasors $\bm{x}_t$ from  $\bm{z}_t = \left[\hat{\boldsymbol{i}}_{\mathcal{A}},\hat{\boldsymbol{v}}_{\mathcal{A}} \right]^\top$ by solving the regularized least square problem:
\begin{equation}
   \min_{\bm{x}_t} \norm{\bm{z}_t  - \bm{H}\bm{x}_t}^2_2+ \mu_1 (\bm{x}_t^H 
\mathbf{S}
\bm{x}_t)\label{RLS}
\end{equation}
where $\mu_1$ is positive. The closed-form solution of \eqref{RLS} is:
\begin{equation}
  \hat{\bm{x}}_t = \left(\bm{H}^H\bm{H}  + \mu_1
\mathbf{S}
\right)^\dagger\bm{H}^H\bm{z}_t, \label{recover_vp}
\end{equation}
where $\hat{\bm{x}}_t$ is the estimated voltage phasor and   $(\cdot)^\dagger$ denotes the pseudo-inverse.
The algorithm step are as follows:
\begin{enumerate}
    \item We collect $T$ historical measurements $\bm{z}_{t-T+1}, \cdots, \bm{z}_{t}$.
    \item We utilize \eqref{recover_vp} to obtain the estimated full observations $\hat{\bf{X}} = [\hat{\bm{x}}_{t-T+1}, \ldots, \hat{\bm{x}}_{t}]$.
    \item The Cplx-STGCN loss function for the voltage phasor prediction is written as
    \begin{align}
     &\mathcal{L}(\Phi, \theta)  = \sum_t \Big\{\norm{\bm{y}^r_t - \bm{x}_{t+H} }^2 +\label{objfore}\\
     & \mu_2 \norm{ \hat{\boldsymbol{v}}_{t+H,\mathcal{A}}  \circ \hat{\boldsymbol{i}}_{t+H, \mathcal{A}}^*  - \big[\bm{y}^r_t \circ  (\mathbf{S} \bm{y}^r_t)^*\big]_{\mathcal{A}}}^2 \Big\},\notag
    \end{align}
    where $\bm{x}_{t+H}$ is the ground truth voltage phasor in the next $H$ time step and $\bm{y}^r_t = \Phi^r(\hat{\mathbf{X}}_t, \mathbf{S}, \theta ) $ in \eqref{predictlabel} is the predicted target to approximate the ground-truth regression target $ (\bm{x}_{t+H})$, and the regularization term in \eqref{objfore} favors voltage phasor forecasts that minimize the sum of the absolute value of the apparent power injections. Note that $H = 0$ is the voltage phasor estimation, and $H \ge 1$ is the voltage phasor forecasting.
\end{enumerate}
After   training,  we use $\Phi^{r*}(\hat{\mathbf{X}}_t, \mathbf{S}, \theta^* )$ to forecast the outputs $\bm{x}_{t+H}$ from the inputs $[\bm{x}_{t-T+1}]_{\mathcal{A}}, \ldots, [\bm{x}_{t}]_{\mathcal{A}}$ that are obtained from the observations at the corresponding times.

\subsection{False Data Detection and Localization}
The conventional task of FDI attack detection is a binary hypothesis testing problem where the null-hypothesis is {\it no false data are present} and the positive one is that {\it some data are compromised}. The localization of the false measurement  (FDI localization problem), is the one of interest in this subsection. Such problem amounts to classifying each measurements into two categories (false data or not) and, thus, is a multi-label classification problem. 

%

In a stealth attack \cite{ramakrishna2021gridgraph}, the attacker manipulates both current and voltage phasor measurements on the subset buses $\mathcal{C}$ by introducing a perturbation:
\begin{equation}
\delta \boldsymbol{x}_{t}^{\top}=\left[\begin{array}{ll}
\delta \boldsymbol{x}_{\mathcal{C}}^{\top} & \mathbf{0}_{|\mathcal{P}|+|\mathcal{U}|}^{\top}
\end{array}\right], \text { such that } \boldsymbol{Y}_{\mathcal{P C}} \delta \boldsymbol{x}_{\mathcal{C}}=\mathbf{0},   \mathcal{C}  \subset \mathcal{A}
\end{equation}
where $\mathcal{P}$ is the set of honest nodes. This requires special conditions and placement, since $\boldsymbol{Y}_{\mathcal{P C}}$ is tall and does not have full column-rank for a sufficient number of attacker $\mathcal{C}$. Therefore, the received data $\bm{z}_t$ with FDI attack have the structure:
\begin{equation}
    \bm{z}_t = \bm{H}(\bm{x}_t + \delta \boldsymbol{x}_{t}) + \boldsymbol{\varepsilon}_{t}.\label{attack} 
\end{equation}

\subsubsection{Methodology}
Then, the algorithm for FDI localization is summarized as
\begin{enumerate}
    \item We obtain $T$ historical measurements $\bm{z}_{t-T+1}, \cdots, \bm{z}_{t}$ with FDI attacks based on \eqref{attack}.
    \item We construct the ground-truth label vector $ {\bm{y}}^\prime_t = \text{logit}(\delta \bm{x}_t) $, where $\text{logit}(\cdot)$ is an indicator function  that $[{\bm{y}^\prime}_t]_i = 1$ if $[\delta\bm{x}_t]_i \neq 0$, otherwise $[ {\bm{y}}^\prime_t]_i = 0$. Note that $\mathcal{C}\subset \mathcal{A}$ is the set of randomly sampled buses with the fixed number.
    \item We utilize \eqref{recover_vp} to obtain the estimated voltage phasors $\hat{\bm{x}}_{t-T+1}, \cdots, \hat{\bm{x}}_{t}$.
    \item The loss function of the Cplx-STGCN function for voltage phasor prediction is written as
    \begin{equation}
     \mathcal{L}(\Phi^c, \theta^c)  = \sum_t \norm{ \Phi^c(\hat{\mathbf{X}}_t, \mathbf{S}, \theta^c ) -  {\bm{y}}^\prime_t }^2.
    \end{equation}
\end{enumerate}
\vspace{-0.3cm}
With the trained $\Phi^{c*}(\hat{\mathbf{X}}_t, \mathbf{S}, \theta^c )$, we could predict the multiple labels $\bm{y}^c_t =  \Phi^{c*}(\hat{\mathbf{X}}_t, \mathbf{S}, \theta^c ) $ when $\bm{z}_{t-T+1}, \cdots, \bm{z}_{t}$ are observed.

\section{Numerical Experiments}
The numerical results in this section are  obtained from the IEEE 118-bus case with 118 nodes and 186 edges \cite{zimmerman2010matpower}. This system includes 54 generators, 118 buses (nodes) and 186 edges, whose system parameters can be found in Matpower \cite{zimmerman2010matpower}.  We collect realistic demand profiles from the  \href{https://www.ercot.com/gridinfo/load/load_hist}{\color{blue}Texas grid} and use Matpower \cite{zimmerman2010matpower} to compute the  voltage phasors. In all simulations, we repeat the training and testing 8 times to report the average values.    
 \begin{table}[!htb]\scriptsize
 \center
 \begin{threeparttable}
 \caption{Sensor Measurement Installed Buses}
 \label{pmu_selec}
 \centering
 \begin{tabular}{c |c  }
 \toprule
 Systems&  Bus Name   \\
 \midrule
 ieee 118-bus  & 14,	117, 72, 86,	43,	67,	99,	87,	16,	33,	112,	28,	98,\\
 &	111,	53,	97,	42,	107,	48,	22,	46,	13,	24,	101,	44,	73,	109,	29,	20,\\
 with PMUs & 	91	26,	84,	10,	1,	52,	57,	76,	115,	39,	74,	104,	93,  79,	35,	6,	\\
  & 18,	88,	60,	116,	55,	58,	68,	64,	7,	50,	103,	75,	78,	83,	69.\\
 \bottomrule
 \end{tabular}
 \end{threeparttable}
 \vspace{-0.2cm}
 \end{table}
\begin{figure*}[htbp]   %
\centering
\begin{minipage}{0.65\textwidth}
    \centering
    \subfigure[MSE: PSSF for Voltage Phasors]{\includegraphics[width=0.48\textwidth]{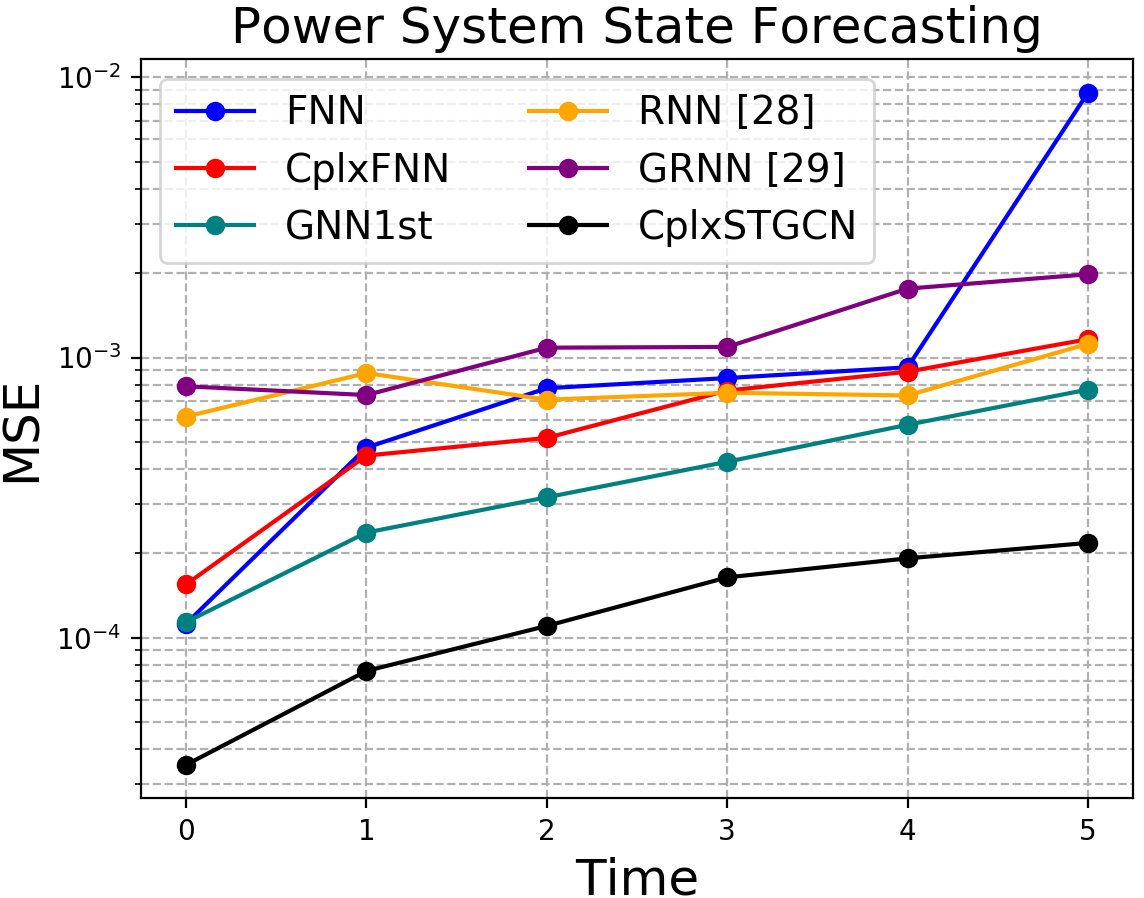}\label{fig1}}
    \hspace{-0.05in}
    \vspace{-0.05in}   
    \subfigure[MSE: PSSF for Voltage Phasors with the GSO change]{\includegraphics[width=0.48\textwidth]{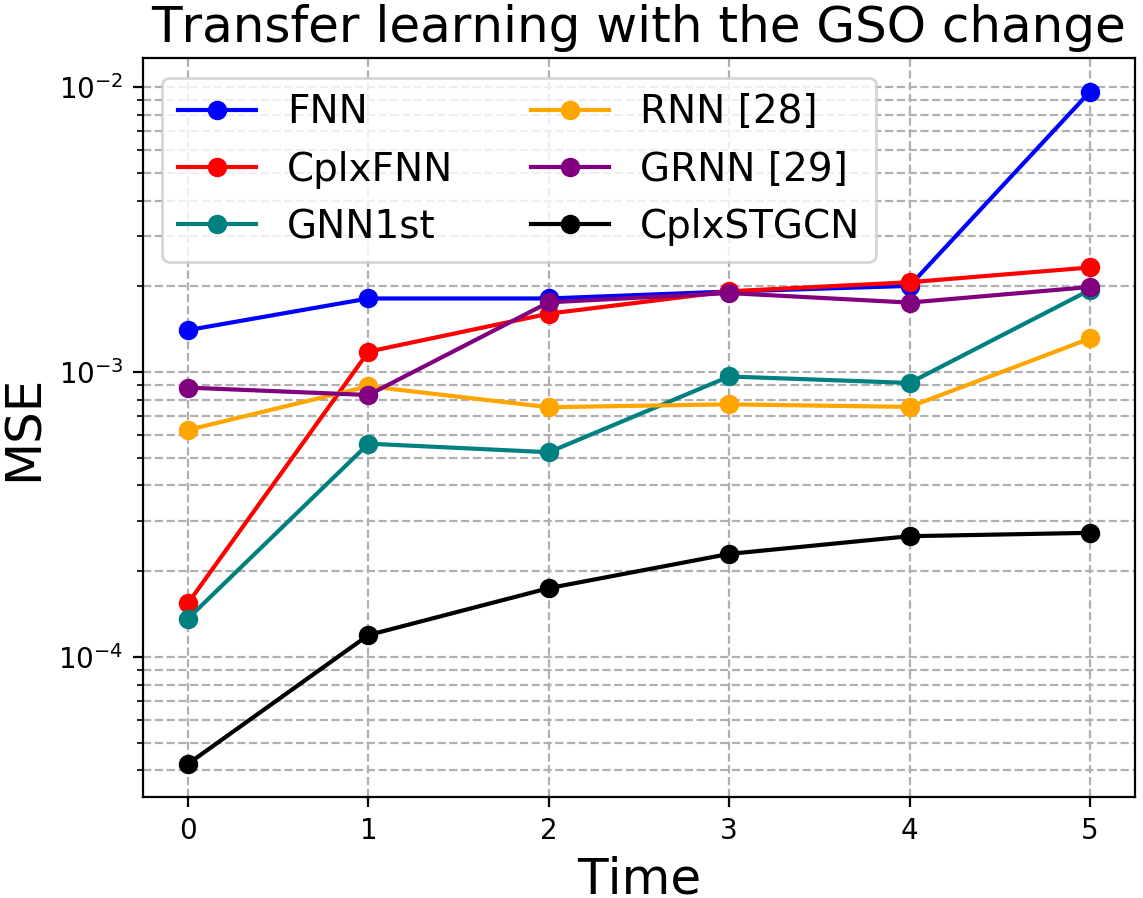}\label{fig2}}
    \hspace{-0.05in}
    \vspace{-0.05in}     
     \subfigure[MAPE: PSSF for Fuel Costs]{\includegraphics[width=0.48\textwidth]{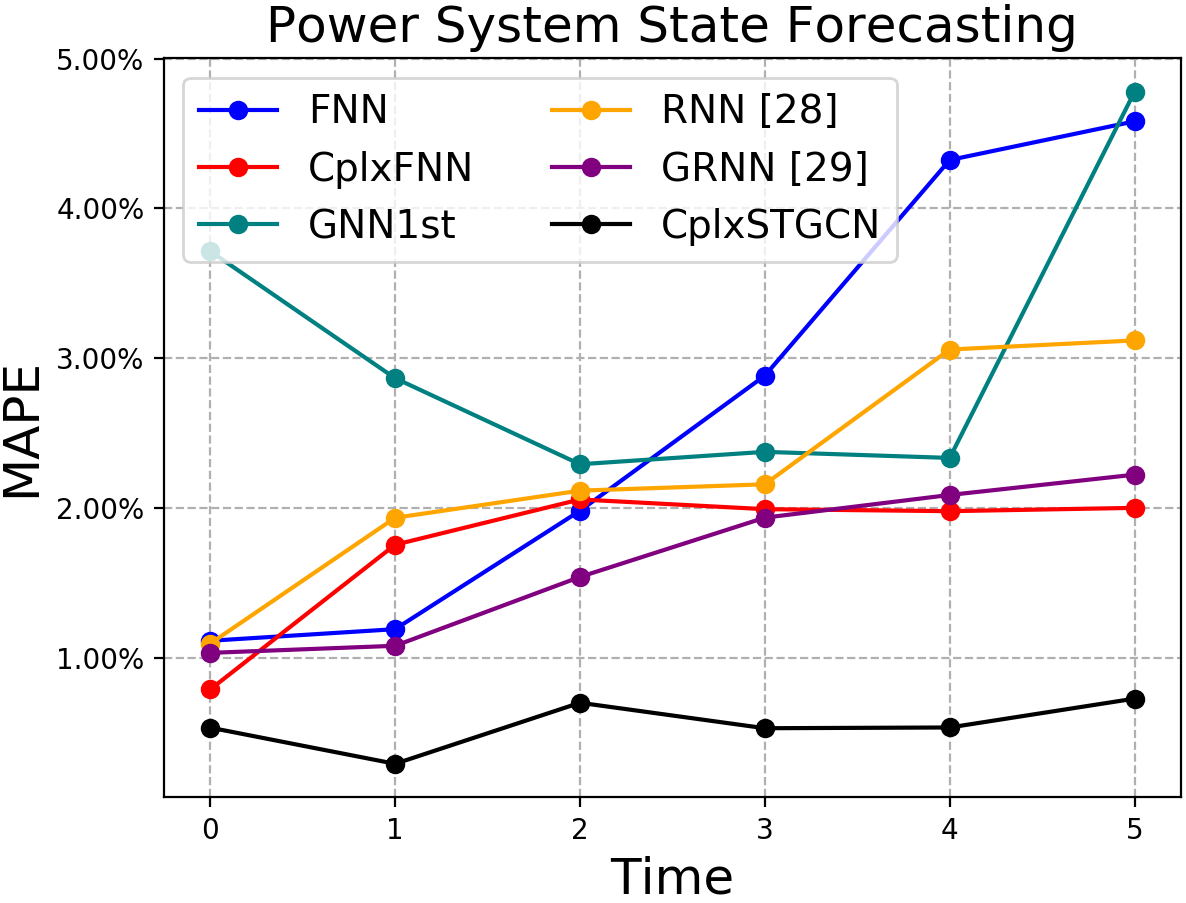}\label{fig3}}
    \hspace{-0.1in}  
    \subfigure[MAPE: PSSF for Fuel Costs with the GSO change]{\includegraphics[width=0.48\textwidth]{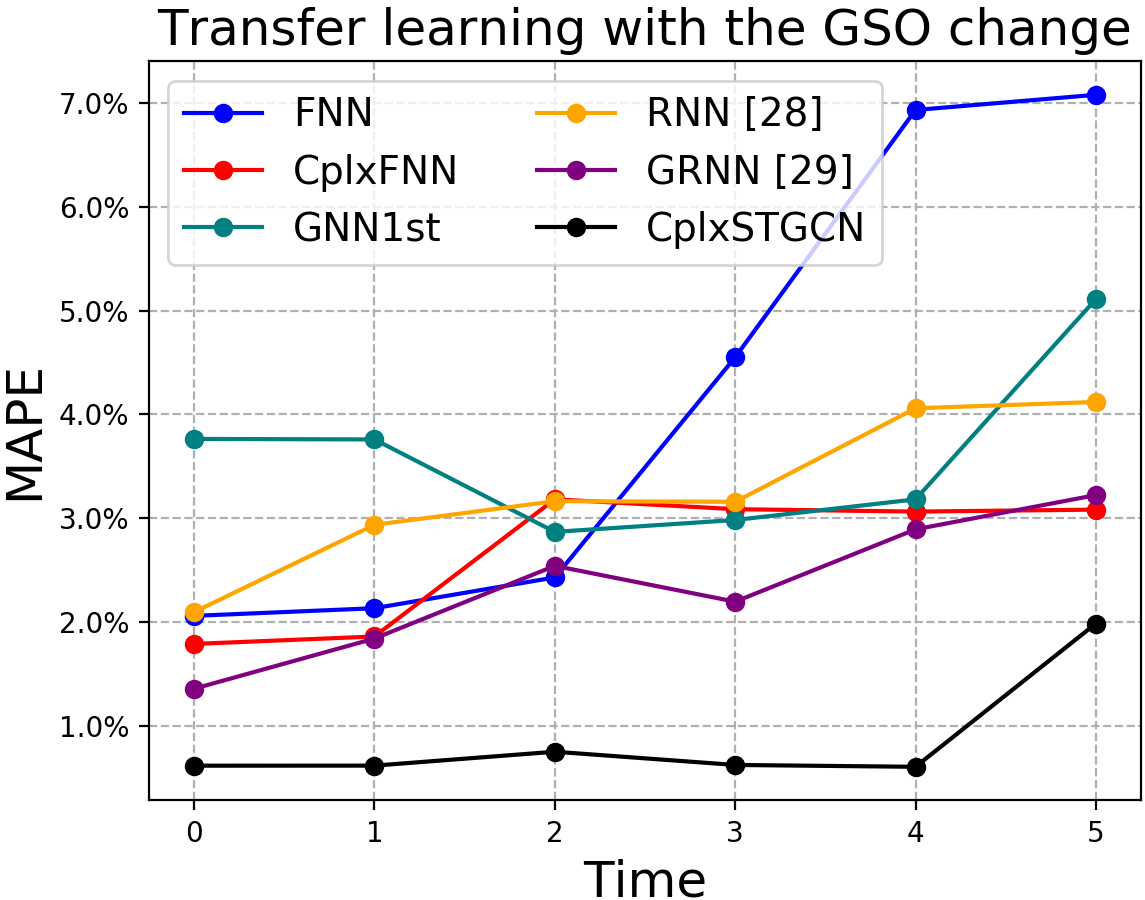}\label{fig4}}
    \caption{Power System State Estimation and Forecasting in the IEEE 118-bus system}
\end{minipage}
\hfill  
\begin{minipage}{0.32\textwidth}
    \centering
    \subfigure[Accuracy: False Data Injection Localization]{\includegraphics[width=1\textwidth]{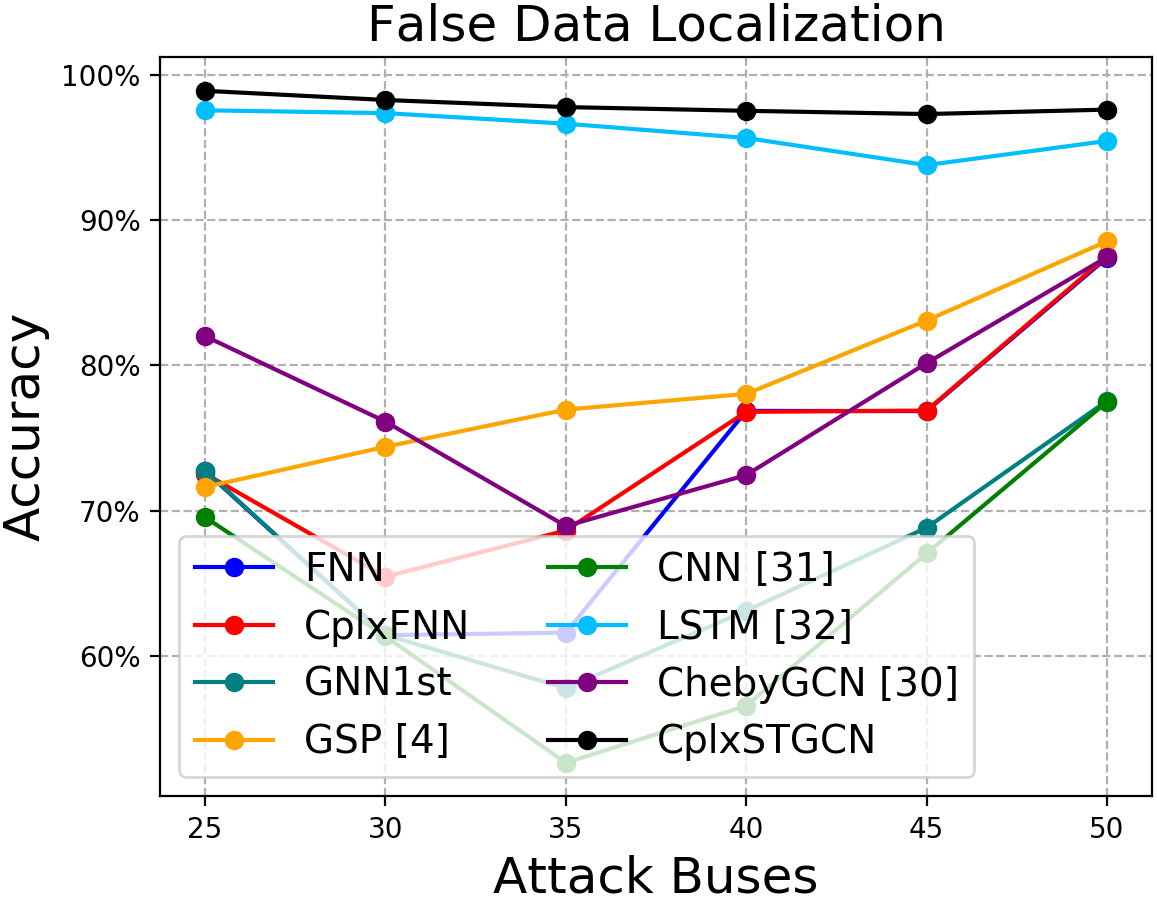}\label{fig5}}
    \hspace{-0.05in} 
    \subfigure[Accuracy: FDI Localization for Hybrid Datasets]{\includegraphics[width=1\textwidth]{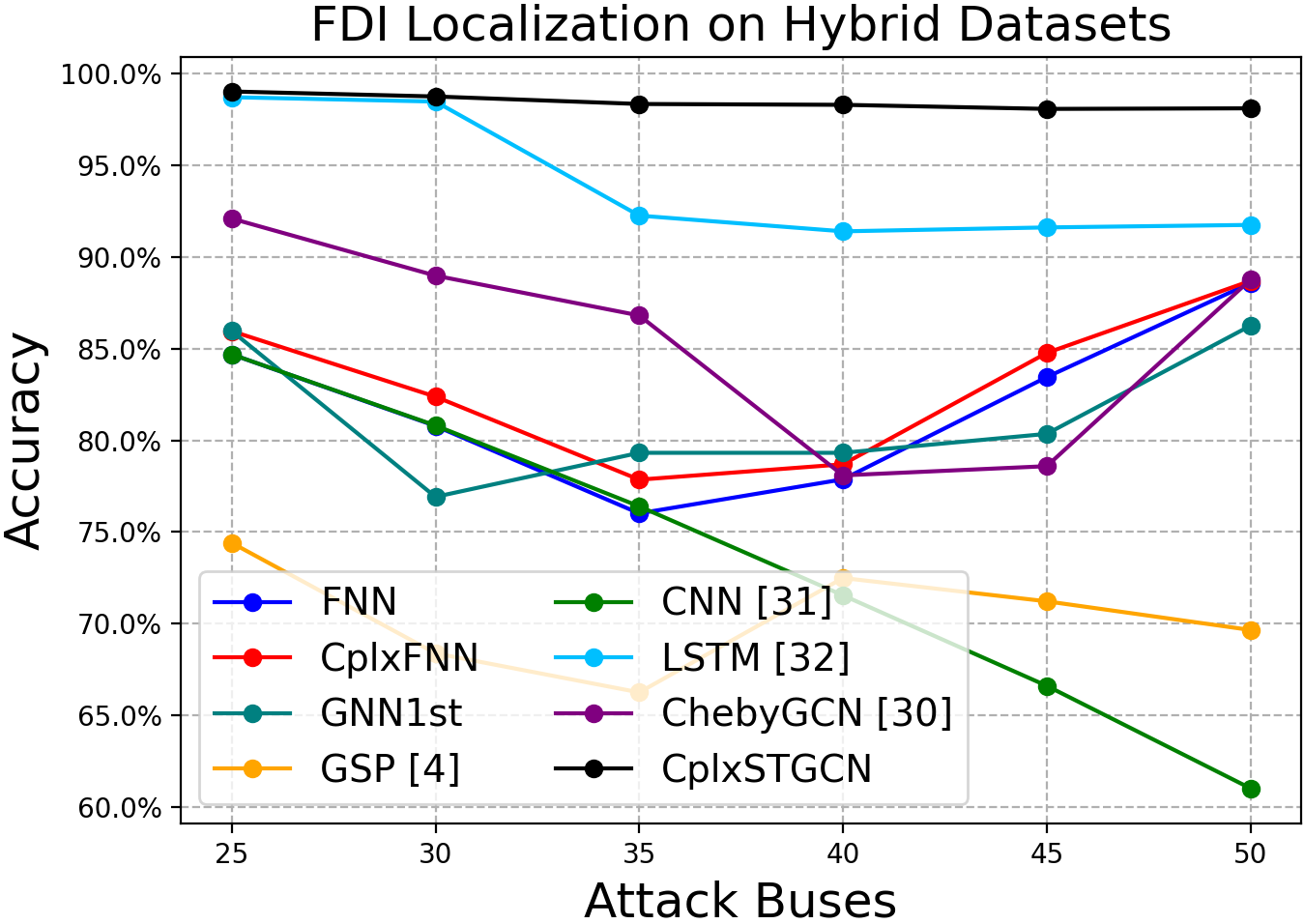}\label{fig6}}
    \caption{False Data Injection Localization in the IEEE 118-bus system.}
\end{minipage}
\vspace{-0.5cm}
\end{figure*}

\textbf{Cplx-STGCN Setting}: The architecture of the Cplx-STGCN for PSSF and FDIA locational detection includes the Cplx-STGCN layer, the two-layer Cplx-NNs and one real-valued output layer. In the Cplx-STGCN layer, the Cplx-CNN (temporal convolution) output channel is 10 and the Cplx-GCN (spatial convolution) output channel is 10. Other Cplx-NNs have 512 neurons per layer. The order of GSO $K$ in the Cplx-GCN is 5. 

\textbf{Baseline Setting}:
The baseline algorithms for both applications include fully-connected NN (\textbf{FNN}) that has 4 layers with 512 neurons each  layers, complex-valued fully-connected NN (\textbf{CplxFNN}) that has 4 layers with 512 neurons each  layers, and \textbf{GCN1st} that has the first layer GNN \cite{kipf2017semi}, and 3 layers of fully connected NNs.
Specific algorithms for PSSF include \textbf{RNN} \cite{zhang2019power} that incorporates a lot of measurements (i.e. voltage magnitudes, active and reactive power injections) and \textbf{GRNN} \cite{hossain2021state} that combines a GNN1st layer and a LSTM layer together to capture the spatio-temporal correlations. Likewise, the state-of-art algorithms for false data detection and localization include \textbf{ChebyGCN} \cite{boyaci2021joint} that uses the absolute values of the admittance matrix and consider active and reactive powers as inputs, \textbf{CNN} \cite{wang2020locational} that takes both line and bus measurements\footnote{For a fair comparision, we choose the same number of sensors as our algorithm},  and \textbf{LSTM} \cite{wang2021kfrnn} that considers voltage phasors as inputs, and \textbf{GSP} algorithm \cite{ramakrishna2021gridgraph}.

\textbf{Sensor Placement}: We take the  eigendecomposition of  $\bm{Y}$ and choose $\abs{\mathcal{K}} = 40$ graph frequency components. Then, we choose the number of sensor placements $\abs{\mathcal{A}} = 60$ and place them so as to maximize $\max_{ {\mathcal{F}}_{\mathcal{A}}}\varpi_{\min} ({\mathcal{F}}_{\mathcal{A}} \mathbf{U}_{\mathcal{K}})$. The resulting of sensor placement is shown in Table \ref{pmu_selec}. Besides, through numerous simulations for the hyperparameter tuning, we choose $\mu_1$ = $1e-6$ and $\mu_2$ = $1e-4$ for all benchmarks.

%

\begin{figure}[!htb]
  \centering
  \subfigure[Voltage Magnitudes ($H=0$).]{
 \includegraphics[width=1.6in]{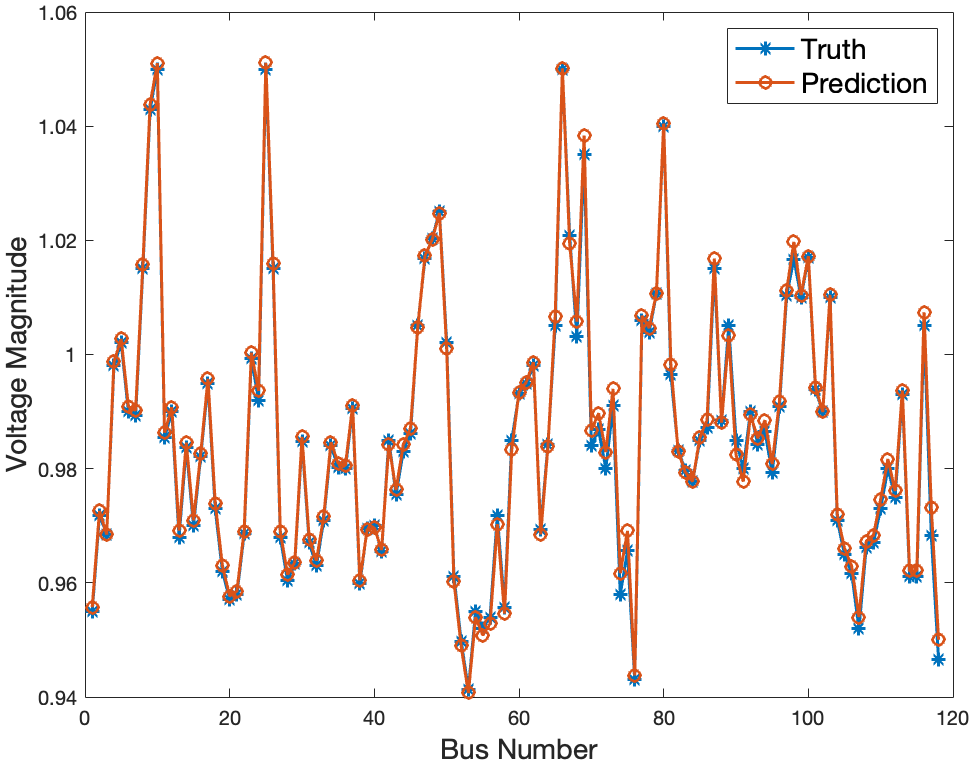}
      \label{Figure1a}
 } 
 \subfigure[Voltage Phase Angles ($H=0$).]{
\includegraphics[width=1.6in]{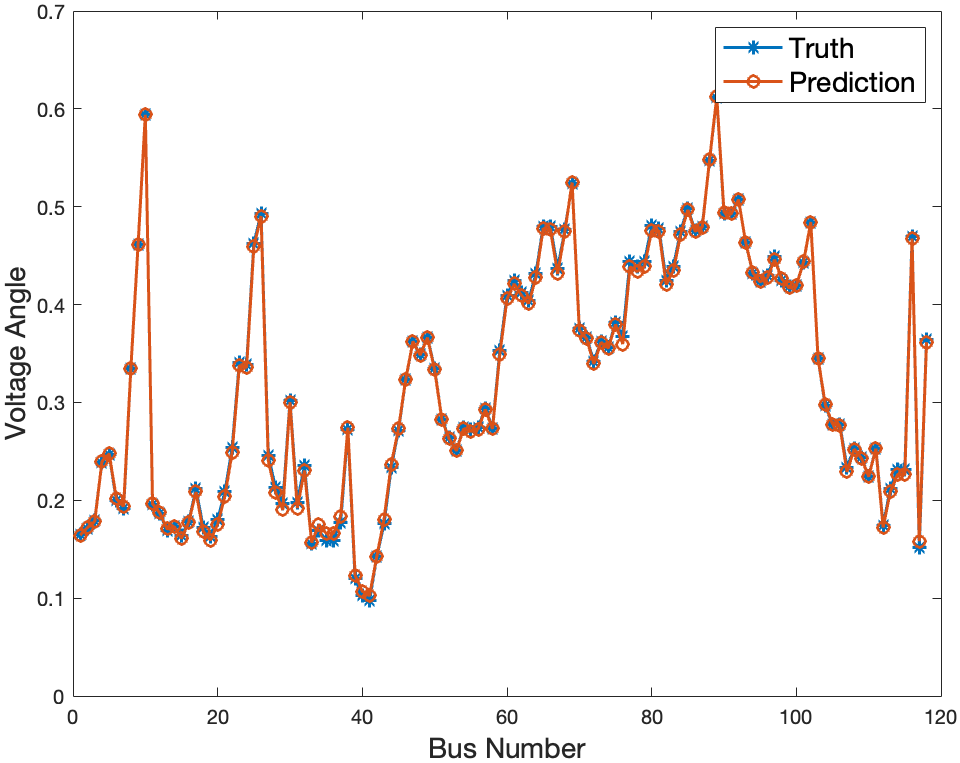}
     \label{Figure1a}
} 
  \subfigure[Voltage Magnitudes ($H=1$).]{
 \includegraphics[width=1.6in]{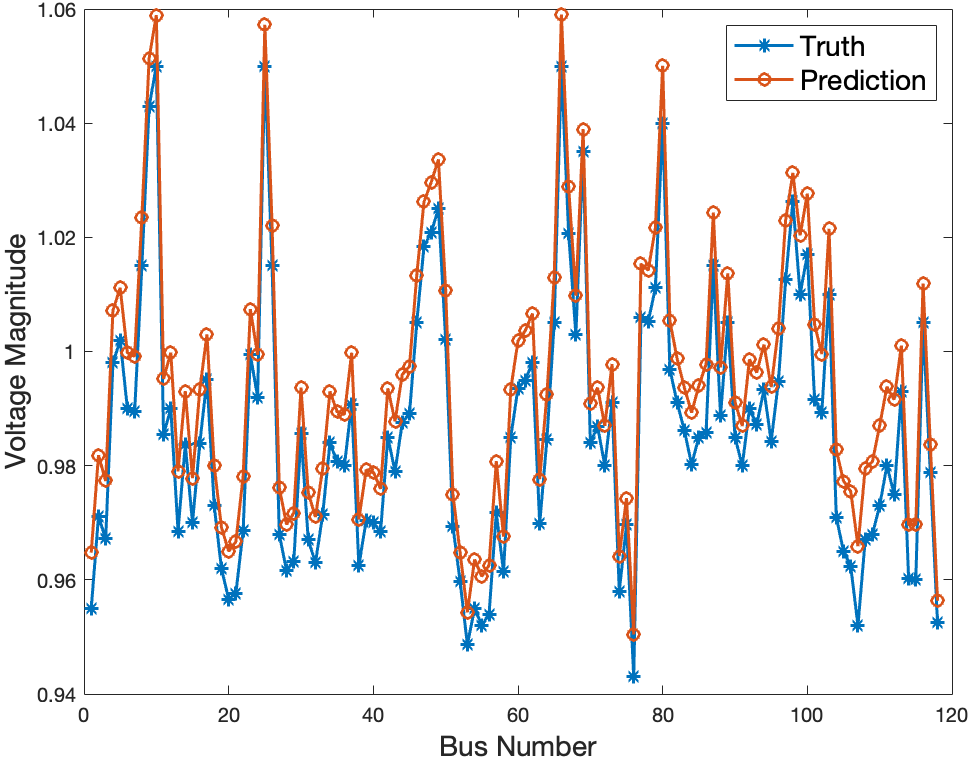}
      \label{Figure1a}
 } 
 \subfigure[Voltage Phase Angles ($H=1$).]{
\includegraphics[width=1.6in]{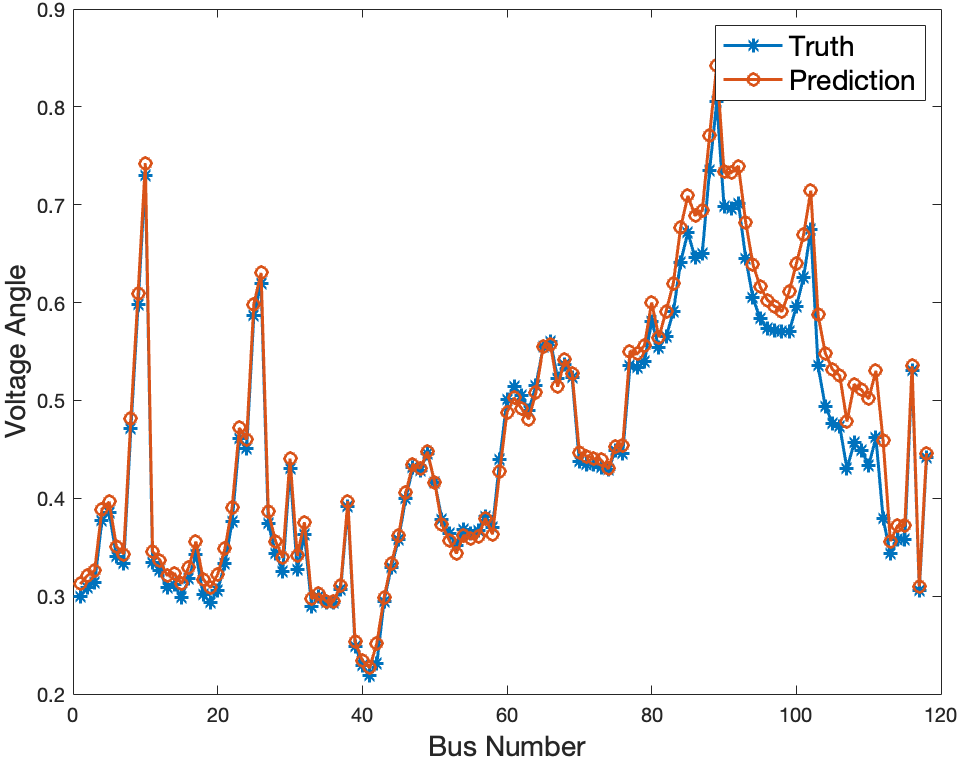}
     \label{Figure1a}
} 
  \caption{An example of PSSE and forecasting.}
  \vspace{-0.4cm}\label{estimation_forecast}
\end{figure}
\subsection{Power System State Estimation and Forecasting}
\subsubsection{Power System State Forecasting Setting}  All the tests use 10 hours as the historical time window, a.k.a. 10 observed data points $T=10$ are used to forecast voltage phasors in the next 1, 2, 3, 4, and 5 hours ($H= 1, 2, 3, 4 , 5$). If $H=0$, it is a PSSE problem that estimates the complete voltage phasors $\bm{x}_{t}$ given $[\bm{x}_{t-T+1}]_{\mathcal{A}}, \ldots, [\bm{x}_{t}]_{\mathcal{A}}$.
\subsubsection{Results}
Figs. \ref{fig1} and \ref{fig3} show the results of Cplx-STGCN and various baselines on the IEEE 118-bus system experiment described above. The Cplx-STGCN achieves the best performance. In particular,  $H = 0$ is the PSSE problem, and the MSE of \eqref{recover_vp} for estimation  is 0.0002371. While this is a respectable outcome, the supervised Cplx-STGCN has much smaller error, i.e. 0.00003517. Another observation is that the voltage phasors predicted by Cplx-STGCN could approximate the OPF results with much smaller MAPE, e.g. 0.5359\% at $H = 4$, compared with other methods, e.g. 2.0875\% of GRNN and 4.3227\% of FNN.
 We illustrate two examples in Fig. \ref{estimation_forecast} to show the ground-truth fully-observed voltage  magnitudes and phase angles with the predicted  ones with $H=0, 1$, which shows the predicted  voltage phasors are very close to the ground-truth. 
\subsubsection{Transferability of Cplx-STGCN Regarding Topology Changes}
In this section, we validate the transferability of the proposed Cplx-STGCN against the topology changes. Retraining a new model based on the new topology is time-consuming.  To handle this problem, we  keep the trained parameters unchanged and modify the GSO of Cplx-STGCN corresponding to  the topology changes of power grids. In this simulation, we trip one line of power grids as the new topology. The results are shown in Fig. \ref{fig2} and \ref{fig4}, which indicates that Cplx-STGCN performs well in the new topology. However, the fully-connected neural networks do not adapt to  the new topology well. This is because the GSO captures the topology changes while the fully-connected neural networks do not have this property. 
%
%
\vspace{-0.4cm}

\subsection{False Data Injection  Localization}
\subsubsection{FDI Setting}  The output of Cplx-STGCN $\bm{y}_t \in [0, 1]^{\abs{\mathcal{A}}}$ can be classified by a discrimination threshold (i.e. 0.5) to quantify the outputs to 0 or 1. 
The discrimination threshold can be adjusted to increase or decrease the sensitivity to application factors. Unless specified,  the discrimination threshold is set to 0.5 in this article following the common practice. 
 Likewise, 10 observed data points $T=10$ are used for FDI localization. 
The number of measurements $\abs{\mathcal{A}}$ is 60, and  the 60 binary labels $\bm{y}_t = [y_1, y_2, \cdots, y_{60} ]^\top$ are converted into one label with a class size of $2^{60}$. Note that, $\abs{\mathcal{C}}$ denotes the number of buses attacked, which is chosen from 25 to 50. 
We also provide the accuracy, \textbf{precision}, \textbf{recall}, and \textbf{$F_1$ scores} with biases  for the Cplx-STGCN. 

\subsubsection{Results}  The performance comparison for FDI localization is  in Fig. \ref{fig5}. We can observe that Cplx-STGCN has much higher accuracy over other NNs and the GSP method that solves a LASSO problem to detect the false data entries. Other NNs tend to predict all ones or all zero vectors depending on whether $\abs{\mathcal{C}}$ is large or small, respectively, while Cplx-STGCN   captures the high-order spatial dependency and temporal correlation of voltage phasors, and thus achieve better performance as a result. Another observation is Cplx-STGCN has higher accuracy than LSTM \cite{wang2021kfrnn}, especially when $\abs{\mathcal{C}}$ is large. Finally, Cplx-STGCN exhibits more stable performance with different values $\abs{\mathcal{C}}$.

\subsubsection{FDI Localization for Hybrid Dataset}
In the previous subsection, we have considered the attack hypotheses that for every $\bm{z}_t, \forall t$, FDI attacks are launched on some buses $\mathcal{C}$ of $\bm{z}_t$. In this subsection, we test the proposed algorithm on the data set under both no-attack ($H_0$) and attack ($H_1$)  hypotheses. Therefore, the received data $\bm{z}_t$ with FDI attack have the structure:
\begin{equation}
\boldsymbol{z}_{t}= \begin{cases}H_{0}: \bm{H}(\bm{x}_t ) + \boldsymbol{\varepsilon}_{t} \\ H_{1}:  \bm{H}(\bm{x}_t + \delta \boldsymbol{x}_{t}) + \boldsymbol{\varepsilon}_{t}\end{cases}.
\end{equation}
Fig.\ref{fig6} shows the simulation results for the data set under both no-attack ($H_0$) and attack ($H_1$)  hypotheses. It shows that the proposed algorithm has very high accuracy, e.g. 98.1043\%, compared with other methods, e.g. 88.5806\% of FNN and 91.7466\% of LSTM with $\abs{\mathcal{C}} =50$.

\begin{table}[!htb]
 \renewcommand\tabcolsep{3.0pt}
\scriptsize
\center
\begin{threeparttable}
\caption{FDI Localization Metrics}
\label{FDIlocal_metrics}
\centering
\begin{tabular}{c |c  c  c  c  c  c}
\toprule
Metrics&  $\abs{\mathcal{C}} = 25$ &  $\abs{\mathcal{C}} = 30$ &  $\abs{\mathcal{C}} = 35$ &  $\abs{\mathcal{C}} = 40$ &  $\abs{\mathcal{C}} = 45$ &  $\abs{\mathcal{C}} = 50$  \\
\midrule
Accuracy &  98.8825\%   &	98.2500\%   &	97.7617\%   & 97.5067\%   &	97.2827\%   &	97.5883\%  \\
Precision &    	96.2615\%   &  96.0518\%  &	 96.0790\%  &	 96.1384\%  &	 96.8030\%  &	97.5457\%  \\
Recall &    99.6952\%   &  99.5566\%  &	 99.1962\%  &	 99.8886\%   &	 98.7180\%  &	99.0236\%  \\
$F_1$ Score & 97.9426\%   &  97.7680\%  &	 97.6091\%  &	97.9752\%  &	97.7490\%  &	98.2778\%  \\
\bottomrule
\end{tabular}
\end{threeparttable}
\vspace{-0.1cm}
\end{table}
\begin{table}[!htb]
 \renewcommand\tabcolsep{3.0pt}
\scriptsize
\center
\begin{threeparttable}
\caption{FDI Localization Metrics for Hybrid Datasets}
\label{FDIlocal_other}
\centering
\begin{tabular}{c |c  c  c  c  c  c}
\toprule
Metrics&  $\abs{\mathcal{C}} = 25$ &  $\abs{\mathcal{C}} = 30$ &  $\abs{\mathcal{C}} = 35$ &  $\abs{\mathcal{C}} = 40$ &  $\abs{\mathcal{C}} = 45$ &  $\abs{\mathcal{C}} = 50$  \\
\midrule
Accuracy &    	99.0230\%   &  98.7502\%  &	 98.3413\%  &	 98.2958\%  &	 98.0729\%  &	98.1043\%  \\
Precision &    	94.2437\%   &  94.4576\%  &	 94.3624\%  &	 95.5202\%  &	 95.8494\%  & 96.9704\%  \\
Recall &    99.1154\%   &  99.0222\%  &	 98.2522\%  &	 98.4301\%  &	 97.7849\%  &	98.0426\%  \\
$F_1$ Score &  96.5963\%   &  96.6700\%  &	 96.2546\%  &	 96.9452\%  &	 96.8011\%  &	97.4988\%  \\
\bottomrule
\end{tabular}
\end{threeparttable}
\vspace{-0.2cm}
\end{table}
\subsubsection{Other Metrics of FDI Localization}
In this section, we show other metrics, including  precision, recall and $F_1$ score.   Let  True Negatives ($TN$) refer to the unattacked buses that are classified as unattacked buses. True Positives ($TP$) refer to the FDI attacked bus correctly predicted to be attacked. False Negatives ($FN$) refer to the unattacked buses that are predicted to be attacked, and False Positives ($FP$) refer to the attacked buses  that are predicted to be unattacked.
Moreover, precision is defined as the ratio of the number of $TP$  to the total number of buses that are actually attacked. Likewise, recall (True Positive Ratio (TPR)) is defined as the ratio of the number of $TP$ to the number of true attacked buses
   \begin{align}\small
   	\text{Precision}=\frac{TP}{TP+ FP},  ~~~~~  	\text{Recall}=\frac{TP}{TP+FN}.
   \end{align}
 To strike a balance between the precision and recall, we  define $F_1$-Score as the geometrical average of the precision and recall.  Eq. \eqref{f1_score} shows the $F_1$-Score calculation. 
      \begin{align}\small
   	F_1\text{-Score}=2\times \frac{\text{Precision}\times \text{Recall}}{\text{Precision}+ \text{Recall}}.\label{f1_score}
   \end{align} 
The results are shown in Tables \ref{FDIlocal_metrics} and \ref{FDIlocal_other}. It shows that Precision, Recall and $F_1$ Score are very high. It indicates that the false alarm rate of the Cplx-STGCN is  small and  for both balanced and unbalanced data sets.



\section{Conclusions}
In this paper, we presented a new complex-valued spatio-temporal graph convolutional neural network architecture for the complex-valued graph signals and graph shift operator.  Among the key advantages of our approach compared to traditional methods is its more informative and compact representation for complex-valued GS. We show two applications of the proposed Cplx-STGCN in power systems that have the complex-valued GSO and GS, including power system state forecasting and FDI localization.  We prove that complex-valued GCNs are stable with respect to perturbations of the underlying graph support and both  the transfer error and the    propagation error through multiply layers are bounded. The results of the experiments attest the potential of the nascent field of geometric deep learning in the complex domain, and can spur future research in Artificial Intelligence for energy system, wireless communication and biological networks whose signals are sparse in the Fourier domain. 

\appendix





\subsection{Proof of Theorem 1}\label{app:Thm1}
\noindent
\textit{Proof}: We want to characterize how much a change in   $\hat{\mathbf{S}}$ changes the response of the feature extraction layer as
\begin{equation}
\begin{aligned}
& \hat{\mathcal{H}}(\hat{\mathbf{S}}) -  {\mathcal{H}}(\hat{\mathbf{S}}) = \sum_{k=0}^K  (\hat{h}_k -   h_k) \hat{\bf{S}}^k  = \sum_{k=0}^K  (\hat{h}_k -   h_k) (\mathbf{S} +\mathbf{E})^k 
	 \end{aligned}
\end{equation}
We are interested in characterizing how different are the outputs of $\hat{\mathcal{H}}(\hat{\mathbf{S}})$ and ${\mathcal{H}}(\hat{\mathbf{S}})$. 
Let $\sigma_{\max}(\mathbf{A})$ be the largest singular value of  matrix $\bm A$; we know that $\|\bm A\bm x\|\leq \sigma_{\max}(\bm A)\|\bm x\|$. Hence, for a given input, the norm of the difference of the output of the first layer is scaled at most by: 
\begin{equation}
\begin{aligned}
\sigma_{\max}(\hat{\mathcal{H}}(\hat{\mathbf{S}})&\!-\! {\mathcal{H}}(\hat{\mathbf{S}})) = \max_{i} \left|\sigma_i\big( \hat{\mathcal{H}}(\hat{\mathbf{S}})\!-\! {\mathcal{H}}(\hat{\mathbf{S}}) \big) \right|   \\
&= \max_i \bigg| \sum_{k=0}^K (\hat{h}_k\!-\! h_k) \sigma_i^k( \mathbf{S}+\mathbf{E}) \bigg|.
	 \end{aligned}
\end{equation}

 Furthermore, the following inequalities hold:
\begin{equation}
\begin{aligned}
&\max_{\norm{\mathbf{E}} \le \epsilon} \sigma(\hat{\mathcal{H}}(\hat{\mathbf{S}})\!-\! {\mathcal{H}}(\hat{\mathbf{S}})) = \max_{z\in \Sigma (\hat{\mathbf{S}})} \bigg|\sum_{k=0}^K (\hat{h}_k\!-\! h_k) z^k \bigg|\le \\
& \max_{z\in \Sigma (\hat{\mathbf{S}})} \sum_{k=0}^K \abs{\hat{h}_k\!-\! h_k} \abs{z}^k 
=\max_{0\le x \le \sigma_{\max}(\hat{\mathbf{S}})} \sum_{k=0}^K \abs{\hat{h}_k\!-\! h_k} x^k.\label{eqpolymax}
	 \end{aligned}
\end{equation}
where $\Sigma(\hat{\mathbf{S}})$ denotes the set of singular values of $\hat{\mathbf{S}}$.
Note that we can write $\hat{\mathbf{S}}=\mathbf{S}+\mathbf{E}$, and then
Let $\bm v_{\max}$ be the largest right singular vector of $\hat{\mathbf{S}}$; we can write the inequalities:
\begin{equation}
\begin{aligned}
\sigma_{\max}(\hat{\mathbf{S}})&=\max_{\bm x}\frac{\norm{\hat{\mathbf{S}}\bm x} }{\norm{\bm x}}   =   \norm{ ({\mathbf{S}} + {\mathbf{E}})\bm v_{\max}} \\ 
&\le \norm{\mathbf{S}\bm v_{\max}} +\norm{\mathbf{E}\bm v_{\max}}\le \sigma_{\max}( \mathbf{S}) +\epsilon,
\end{aligned}
\end{equation}
Therefore, \eqref{eqpolymax} can be further bounded as:
\begin{equation}
\begin{aligned}
\max_{0\le x \le \sigma_{\max}(\hat{\mathbf{S}})} \sum_{k=0}^K \abs{\hat{h}_k\!-\! h_k} x^k \le \!\!\!\!\max_{0\le x \le \sigma_{\max}( \mathbf{S}) +\epsilon } \sum_{k=0}^K \abs{\hat{h}_k\!-\! h_k} x^k 
\end{aligned}
\end{equation}
This inequality holds due to the fact that the feasible set is relaxed.
Moreover,   the polynomial $ \sum_{k=0}^K \abs{\hat{h}_k\!-\! h_k} x^k$ has all positive coefficients, and thus is bounded by 
\begin{equation}
\begin{aligned}
&\max_{0\le x \le \sigma_{\max}( \mathbf{S}) +\epsilon } \sum_{k=0}^K    \abs{\hat{h}_k\!-\! h_k} x^k \\
& \le   \sum_{k=0}^K    \abs{\hat{h}_k\!-\! h_k} (\sigma_{\max}(\mathbf{S}) +\epsilon )^k\\
&\le  \max\bigg(1, (\sigma_{\max}(\mathbf{S}) +\epsilon )^K  \bigg) \sum_{k=0}^K    \abs{\hat{h}_k\!-\! h_k}\\
& =\overbrace{\max\Big(1, (\sigma_{\max}(\mathbf{S}) +\epsilon )^K  \Big)}^{\gamma_1} \|\hat{\bm h}\!-\! \bm{h}\|_1
\end{aligned}
\end{equation}

Therefore, we can conclude that 
 \begin{equation}
\begin{aligned}
\sigma_{\max}(\hat{\mathcal{H}}(\hat{\mathbf{S}})&\!-\! {\mathcal{H}}(\hat{\mathbf{S}}))  \le \gamma_1 \|\hat{\bm h}\!-\! \bm{h}\|_1
\end{aligned}
\end{equation}
This completes the proof.
\hfill $\blacksquare$
%


\subsection{Proof of Theorem 2}\label{app:Thm2}

\textit{proof}:  
To bound of $\rho({\mathcal{H}}( \hat{\mathbf{S}}) - \mathcal{H}( {\mathbf{S}}) )$, let $\mathbf{E} \triangleq \epsilon \bar{\mathbf{E}}$ and study the following expansion: 
\begin{equation}
\begin{aligned}\label{polyno1}
	 &{\mathcal{H}}( \hat{\mathbf{S}}) - \mathcal{H}( {\mathbf{S}})  = \sum_{k=0}^K h_k[(\mathbf{S} + \mathbf{E})^k -  \mathbf{S}^k] \\
	 & = \sum_{k=1}^K h_k  \sum_{\ell=1}^k {k \choose \ell}   {\mathbf{E}}^\ell \mathbf{S}^{k-\ell}  
	 = \sum_{k=1}^K h_k  \sum_{\ell=1}^k {k \choose \ell}  \bar{\mathbf{E}}^\ell \mathbf{S}^{k-\ell} \epsilon^\ell,  
\end{aligned}
\end{equation}
where it is easily to verify $\norm{\bar{\mathbf{E}}}\le 1$ due to $\norm{{\mathbf{E}}}\le \epsilon$.
We can rewrite \eqref{polyno1} as the following matrix polynomial function:
\begin{equation}
\begin{aligned}\label{polyneps}
	 {P}(\epsilon) = \bm A_1\epsilon + \cdots + \bm A_K\epsilon^K  
\end{aligned}
\end{equation}
%
where the coefficients $\bm A_\ell, \forall \ell = 1, \cdots, K$  are expressed as follows:
\begin{equation}
\begin{aligned}
	 \bm A_\ell = \sum_{k=\ell}^K h_k{k\choose \ell}\bar{\mathbf{E}}^{\ell} \mathbf{S}^{k-\ell},
\end{aligned}
\end{equation}
and in particular, we have $A_K = h_K \bar{\mathbf{E}}^K$. Therefore, we have the bound for the norm of $\bm A_\ell$:
\begin{equation}
\begin{aligned}
	 \norm{\bm A_\ell}_2 & \le  \sum_{k=\ell}^K \abs{h_k}{k\choose \ell}  \norm{\mathbf{S}^{k-\ell}}_2\le \sum_{k=\ell}^K \abs{h_k}{k\choose \ell}  \norm{\mathbf{S}}_2^{k-\ell}\\
	 & =\sum_{k=\ell}^K \abs{h_k}{k\choose \ell}   \sigma_{\max}^{k-\ell}(\mathbf{S})\le \|\bm A\|_1\\
	 &\le
	 \max_{1\leq k\leq K}|h_k|  
	 \sum_{k=1}^K
	 {k\choose \ell}   \sigma_{\max}^{k-\ell}\\
	 &\leq  \max_{1\leq k\leq K}|h_k| (1+\sigma_{\max}(\mathbf{S}))^K
	 \label{coeff1}
\end{aligned}
\end{equation}
Consider the definition \eqref{coeff2} $\gamma_2\triangleq 
\max_{1\leq k\leq K}|h_k| (1+\sigma_{\max}(\mathbf{S}))^K
$.
We have that:
\begin{equation}
\begin{aligned}\label{polyneps}
	 \norm{{P}(\epsilon)}_2 \le  \norm{\bm A_1}{\epsilon} + \cdots + \norm{\bm A_K}{\epsilon}^K  \le \gamma_2 \frac{{\epsilon}(1-{\epsilon}^K)}{1- {\epsilon}}
\end{aligned}
\end{equation}

This completes the proof.
\hfill $\blacksquare$

\begin{lemma}\label{lemma6}
Considering the nonlinear activation function $\operatorname{CReLU}(\cdot)$,  the distance between $\operatorname{CReLU}(\mathcal{H}(\mathbf{S})\bm{x})$ and $\operatorname{CReLU}(\hat{\mathcal{H}}(\hat{\mathbf{S}})\bm{x})$ is also bounded by: 
	\begin{equation}
	\begin{aligned}
	\|\big(
&\operatorname{CReLU} 	(\hat{\mathcal{H}}(\hat{\mathbf{S}})\bm x) -\operatorname{CReLU} 	(\mathcal{H}(\mathbf{S})\bm x) \big)
	\|\le\\ &\left(
	\gamma_1
\left\|\hat{\bm{h}}- {\bm{h}} \right\|_1  +
\gamma_2 \frac{{\epsilon}(1-{\epsilon}^K)}{1- {\epsilon}}
	\right)\|\bm x\|
	\end{aligned}
\end{equation}
\end{lemma}
\textit{Proof}: Consider two complex numbers $z_1 = x_1 +\mathfrak{j} y_1$ and $z_2 = x_2 +\mathfrak{j} y_2$, the following relationship holds:
\begin{equation}
\begin{aligned}
	&\abs{\operatorname{CReLU}(z_1) - \operatorname{CReLU}(z_2)} = 	\\
	&\abs{\operatorname{CReLU}(x_1+\mathfrak{j} y_1) - \operatorname{CReLU}(x_2+\mathfrak{j} y_2)} =\\
	&  \abs{\operatorname{ReLU}(x_1 - x_2) + \mathfrak{j}\operatorname{ReLU}(y_1 -  y_2)} = \\
	& \sqrt{(\operatorname{ReLU}(x_1 - x_2))^2 + (\operatorname{ReLU}(y_1 -  y_2))^2}\le\\
	& \sqrt{(x_1 - x_2)^2 + (y_1 -  y_2)^2} = \abs{z_1-z_2}.
\end{aligned}
\end{equation}
The inequality follows from $\abs{\operatorname{ReLU}(x_1 - x_2)} \le \abs{x_1 - x_2}$. Then, we consider the distance between $\operatorname{CReLU}(\mathcal{H}(\mathbf{S})\bm{x})$ and $\operatorname{CReLU}(\hat{\mathcal{H}}(\hat{\mathbf{S}})\bm{x})$:
 \begin{equation}
\begin{aligned}
	&\norm{\operatorname{CReLU}(\mathcal{H}(\mathbf{S})\bm{x}_t) - \operatorname{CReLU}(\hat{\mathcal{H}}(\hat{\mathbf{S}})\bm{x})} \le \\ 	
	& \norm{\mathcal{H}(\mathbf{S})\bm{x} - \mathcal{H}(\hat{\mathbf{S}})\bm{x}} \\
	& =  \norm{\mathcal{H}(\mathbf{S})  - \mathcal{H}(\hat{\mathbf{S}})} \norm{\bm{x}}\le \norm{\mathcal{H}(\mathbf{S}) - \mathcal{H}(\hat{\mathbf{S}})}\\
	& \le \left( \gamma_1\left\|\hat{\bm{h}}- {\bm{h}} \right\|_1  +\gamma_2 \frac{{\epsilon}(1-{\epsilon}^K)}{1- {\epsilon}}\right) \norm{\bm x}.
\end{aligned}
\end{equation}
This completes the proof.
\hfill $\blacksquare$

Likewise, we can easily verify that 
\begin{equation}
\begin{aligned}
	&\abs{\operatorname{tanh}(z_1) - \operatorname{tanh}(z_2)} = 	\\
	&\abs{\operatorname{tanh}(x_1+\mathfrak{j} y_1) - \operatorname{tanh}(x_2+\mathfrak{j} y_2)} =\\
	&  \abs{\operatorname{tanh}(x_1 - x_2) + \mathfrak{j}\operatorname{tanh}(y_1 -  y_2)} = \\
	& \sqrt{(\operatorname{tanh}(x_1 - x_2))^2 + (\operatorname{tanh}(y_1 -  y_2))^2}\le\\
	& \sqrt{\abs{x_1 - x_2}^2 + \abs{y_1 -  y_2}^2} = \abs{z_1-z_2}.
\end{aligned}
\end{equation}
The last equation holds due to the 1-Lipschitz property of $\operatorname{tanh}(\cdot)$.

\subsection{Proof of Lemma 1}\label{app:Thm3}
\textit{Proof}:  We could express the multilayer neural networks, i.e., $\bm{y} = \Phi(\bm{x}, \mathbf{S}, \theta)$,  as a function composition:
 \begin{equation}
\begin{aligned}
	& g(\bm{x}) = \mathcal{H}({\mathbf{S}})\bm{x} = \sum_{k= 0}^K   h_k   {\bf{S}}^k \bm{x}, ~~ \bm{x}^1 = \operatorname{CReLU}(g(\bm{x})), \\
	&  f(\bm{x}^1) = \Theta^{cplx} \bm{x}^1, ~~ \bm{y}  =    \operatorname{tanh}(f(\bm{x}^1)).
\end{aligned}
 \end{equation}
Therefore, we could have the following inequality:
 \begin{equation}
\begin{aligned}
	&\norm{{\bm{y}} - \hat{\bm{y}}} =  \norm{ \operatorname{tanh}(f(\bm{x}^1)) -  \operatorname{tanh}(\hat{f}(\bm{x}^1))}\notag \\
	& = \norm{\Theta^{cplx} \bm{x}^1 - \hat{\Theta}^{cplx}\hat{\bm{x}}^1} \notag \\
	&=\norm{\Theta^{cplx} \bm{x}^1 - \Theta^{cplx} \hat{\bm{x}}^1  +  {\Theta}^{cplx}\hat{\bm{x}}^1 - \hat{\Theta}^{cplx} \hat{\bm{x}}^1}\\
	&\le \norm{\Theta^{cplx} \bm{x}^1 - \Theta^{cplx} \hat{\bm{x}}^1} + \norm{\Theta^{cplx} \hat{\bm{x}}^1 - \hat{\Theta}^{cplx}  \hat{\bm{x}}^1}.
\end{aligned}
 \end{equation}
For the first part, we have 
 \begin{equation}
\begin{aligned}
	&\norm{\Theta^{cplx} \bm{x}^1 - \Theta^{cplx} \hat{\bm{x}}^1} \le \norm{\Theta^{cplx}} \\
	&\norm{\operatorname{CReLU}(\mathcal{H}(\mathbf{S})\bm{x}) - \operatorname{CReLU}(\hat{\mathcal{H}}(\hat{\mathbf{S}})\bm{x})} \le  \sigma_{\max}(\Theta^{cplx})  \\
	&  \left[ 	\gamma_1\left\|\hat{\bm{h}}- {\bm{h}} \right\|_1    + \gamma_2 \frac{{\epsilon}(1-{\epsilon}^K)}{1- {\epsilon}}\right] \norm{\bm{x}}.\label{2layer1bound}
\end{aligned}
 \end{equation}
 The last inequality is due to Lemma \ref{lemma6}.
For the second part, we have
 \begin{equation}
\begin{aligned}\label{secondpart}
&	\norm{\Theta^{cplx} \hat{\bm{x}}^1 - \hat{\Theta}^{cplx} \hat{\bm{x}}^1} \le \norm{\Theta^{cplx} - \hat{\Theta}^{cplx}}   \\
& \norm{\operatorname{CReLU}(\hat{\mathcal{H}}(\hat{\mathbf{S}})\bm{x})}\le \delta_{\mathbf{w}}  \norm{ \sum_{k= 0}^K \hat{h}_k  \hat{\bf{S}}^k  } \norm{\bm{x}} \\
& = \delta_{\mathbf{w}}  \norm{ \sum_{k= 0}^K \hat{h}_k   (\mathbf{S} + \mathbf{E})^k  } \norm{\bm{x}}.
\end{aligned}
 \end{equation}
Similar to Theorem \ref{transferbound}, we could bound $\norm{ \sum_{k= 0}^K \hat{h}_k   (\mathbf{S} + \mathbf{E})^k  }$ as follows:
\begin{equation}
\begin{aligned}
&\max_{\norm{\mathbf{E}} \le \epsilon} \sigma(\sum_{k= 0}^K \hat{h}_k   (\mathbf{S} + \mathbf{E})^k  ) = \max_{z\in \Sigma(\hat{\mathbf{S}})} \bigg|\sum_{k=0}^K \hat{h}_k  z^k \bigg|\le \\
& \max_{z\in \Sigma(\hat{\mathbf{S}})} \sum_{k=0}^K \abs{\hat{h}_k } \abs{z}^k 
=\max_{0\le x \le \sigma_{\max}(\hat{\mathbf{S}})} \sum_{k=0}^K \abs{\hat{h}_k } x^k \le \gamma_1 \left\|\hat{\bm{h}}  \right\|_1  	 \end{aligned}
\end{equation}
where $\Sigma(\hat{\mathbf{S}})$ denotes the set of singular values of $\hat{\mathbf{S}}$.
Therefore, we have the bound of \eqref{secondpart} as follows:
	\begin{equation}
	\begin{aligned}
	  \delta_{\mathbf{w}}  \norm{ \sum_{k= 0}^K \hat{h}_k   (\mathbf{S} + \mathbf{E})^k  }  \le   \delta_{\mathbf{w}}  \gamma_1 \left\|\hat{\bm{h}}  \right\|_1   .\label{2layer2bound}
		 \end{aligned}
\end{equation}
By adding \eqref{2layer1bound} and \eqref{2layer2bound}, we have the bound for $\norm{{\bm{y}} - \hat{\bm{y}}}$:
	\begin{equation}
	\begin{aligned}
&\norm{{\bm{y}} - \hat{\bm{y}}}\le      \delta_{\mathbf{w}}  \gamma_1 \left\|\hat{\bm{h}}  \right\|_1   \norm{\bm{x}} + \sigma_{\max}(\Theta^{cplx}) \\
& \left[ 	\gamma_1\left\|\hat{\bm{h}}- {\bm{h}} \right\|_1 + \gamma_2 \frac{{\epsilon}(1-{\epsilon}^K)}{1- {\epsilon}}\right] \norm{\bm{x}}
		 \end{aligned}
\end{equation}
This completes the proof.
\hfill $\blacksquare$

\begin{footnotesize}

\end{footnotesize}





\begin{thebibliography}{10}
\providecommand{\url}[1]{#1}
\csname url@samestyle\endcsname
\providecommand{\newblock}{\relax}
\providecommand{\bibinfo}[2]{#2}
\providecommand{\BIBentrySTDinterwordspacing}{\spaceskip=0pt\relax}
\providecommand{\BIBentryALTinterwordstretchfactor}{4}
\providecommand{\BIBentryALTinterwordspacing}{\spaceskip=\fontdimen2\font plus
\BIBentryALTinterwordstretchfactor\fontdimen3\font minus
  \fontdimen4\font\relax}
\providecommand{\BIBforeignlanguage}[2]{{%
\expandafter\ifx\csname l@#1\endcsname\relax
\typeout{** WARNING: IEEEtran.bst: No hyphenation pattern has been}%
\typeout{** loaded for the language `#1'. Using the pattern for}%
\typeout{** the default language instead.}%
\else
\language=\csname l@#1\endcsname
\fi
#2}}
\providecommand{\BIBdecl}{\relax}
\BIBdecl

\bibitem{leung1991complex}
H.~Leung and S.~Haykin, ``The complex backpropagation algorithm,'' \emph{IEEE
  Transactions on signal processing}, vol.~39, no.~9, pp. 2101--2104, 1991.

\bibitem{trabelsi2018deep}
C.~Trabelsi, O.~Bilaniuk, Y.~Zhang, D.~Serdyuk, S.~Subramanian, J.~F. Santos,
  S.~Mehri, N.~Rostamzadeh, Y.~Bengio, and C.~J. Pal, ``Deep complex
  networks,'' in \emph{International Conference on Learning Representations},
  2018.

\bibitem{bassey2021survey}
J.~Bassey, L.~Qian, and X.~Li, ``A survey of complex-valued neural networks,''
  \emph{arXiv preprint arXiv:2101.12249}, 2021.

\bibitem{ramakrishna2021gridgraph}
R.~Ramakrishna and A.~{Scaglione}, ``{Grid-Graph Signal Processing (Grid-GSP):
  A Graph Signal Processing Framework for the Power Grid},'' \emph{IEEE Trans.
  Signal Process.}, 2021.

\bibitem{sandryhaila2013discrete}
A.~Sandryhaila and J.~M. Moura, ``Discrete signal processing on graphs,''
  \emph{IEEE transactions on signal processing}, vol.~61, no.~7, pp.
  1644--1656, 2013.

\bibitem{ortega2018graph}
A.~Ortega, P.~Frossard, J.~Kova{\v{c}}evi{\'c}, J.~M. Moura, and
  P.~Vandergheynst, ``Graph signal processing: Overview, challenges, and
  applications,'' \emph{Proceedings of the IEEE}, vol. 106, no.~5, pp.
  808--828, 2018.

\bibitem{dong2020graph}
X.~Dong, D.~Thanou, L.~Toni, M.~Bronstein, and P.~Frossard, ``Graph signal
  processing for machine learning: A review and new perspectives,'' \emph{IEEE
  Signal processing magazine}, vol.~37, no.~6, pp. 117--127, 2020.

\bibitem{defferrard2016convolutional}
M.~Defferrard, X.~Bresson, and P.~Vandergheynst, ``Convolutional neural
  networks on graphs with fast localized spectral filtering,'' \emph{Advances
  in Neural Information Processing Systems}, vol.~29, 2016.

\bibitem{bronstein2017geometric}
M.~M. Bronstein, J.~Bruna, Y.~LeCun, A.~Szlam, and P.~Vandergheynst,
  ``Geometric deep learning: going beyond euclidean data,'' \emph{IEEE Signal
  Processing Magazine}, vol.~34, no.~4, pp. 18--42, 2017.

\bibitem{ramakrishna2020user}
R.~Ramakrishna, H.-T. Wai, and A.~Scaglione, ``A user guide to low-pass graph
  signal processing and its applications: Tools and applications,'' \emph{IEEE
  Signal Processing Magazine}, vol.~37, no.~6, pp. 74--85, 2020.

\bibitem{adali2011complex}
T.~Adali, P.~J. Schreier, and L.~L. Scharf, ``Complex-valued signal processing:
  The proper way to deal with impropriety,'' \emph{IEEE Transactions on Signal
  Processing}, vol.~59, no.~11, pp. 5101--5125, 2011.

\bibitem{jablonski2017graph}
I.~Jab{\l}o{\'n}ski, ``Graph signal processing in applications to sensor
  networks, smart grids, and smart cities,'' \emph{IEEE Sensors Journal},
  vol.~17, no.~23, pp. 7659--7666, 2017.

\bibitem{sperduti1997supervised}
A.~Sperduti and A.~Starita, ``Supervised neural networks for the classification
  of structures,'' \emph{IEEE Transactions on Neural Networks}, vol.~8, no.~3,
  pp. 714--735, 1997.

\bibitem{bruna2014spectral}
J.~Bruna, W.~Zaremba, A.~Szlam, and Y.~LeCun, ``Spectral networks and deep
  locally connected networks on graphs,'' in \emph{2nd International Conference
  on Learning Representations, ICLR 2014}, 2014.

\bibitem{yan2018spatial}
S.~Yan, Y.~Xiong, and D.~Lin, ``Spatial temporal graph convolutional networks
  for skeleton-based action recognition,'' in \emph{Thirty-second AAAI
  conference on artificial intelligence}, 2018.

\bibitem{yu2018spatio}
B.~Yu, H.~Yin, and Z.~Zhu, ``Spatio-temporal graph convolutional networks: a
  deep learning framework for traffic forecasting,'' in \emph{Proceedings of
  the 27th International Joint Conference on Artificial Intelligence}, 2018,
  pp. 3634--3640.

\bibitem{seo2018structured}
Y.~Seo, M.~Defferrard, P.~Vandergheynst, and X.~Bresson, ``Structured sequence
  modeling with graph convolutional recurrent networks,'' in
  \emph{International conference on neural information processing}.\hskip 1em
  plus 0.5em minus 0.4em\relax Springer, 2018, pp. 362--373.

\bibitem{gama2020stability}
F.~Gama, J.~Bruna, and A.~Ribeiro, ``Stability properties of graph neural
  networks,'' \emph{IEEE Trans. Signal Process.}, vol.~68, pp. 5680--5695,
  2020.

\bibitem{zhou2020graph}
J.~Zhou, G.~Cui, S.~Hu, Z.~Zhang, C.~Yang, Z.~Liu, L.~Wang, C.~Li, and M.~Sun,
  ``Graph neural networks: A review of methods and applications,'' \emph{AI
  Open}, vol.~1, pp. 57--81, 2020.

\bibitem{liao2021review}
W.~Liao, B.~Bak-Jensen, J.~R. Pillai, Y.~Wang, and Y.~Wang, ``A review of graph
  neural networks and their applications in power systems,'' \emph{Journal of
  Modern Power Systems and Clean Energy}, 2021.

\bibitem{chen2019fault}
K.~Chen, J.~Hu, Y.~Zhang, Z.~Yu, and J.~He, ``Fault location in power
  distribution systems via deep graph convolutional networks,'' \emph{IEEE
  Journal on Selected Areas in Communications}, vol.~38, no.~1, pp. 119--131,
  2019.

\bibitem{zamzam2020physics}
A.~S. Zamzam and N.~D. Sidiropoulos, ``Physics-aware neural networks for
  distribution system state estimation,'' \emph{IEEE Transactions on Power
  Systems}, vol.~35, no.~6, pp. 4347--4356, 2020.

\bibitem{cai2021structural}
L.~Cai, Z.~Chen, C.~Luo, J.~Gui, J.~Ni, D.~Li, and H.~Chen, ``Structural
  temporal graph neural networks for anomaly detection in dynamic graphs,'' in
  \emph{Proceedings of the 30th ACM International Conference on Information \&
  Knowledge Management}, 2021, pp. 3747--3756.

\bibitem{ma2021comprehensive}
X.~Ma, J.~Wu, S.~Xue, J.~Yang, C.~Zhou, Q.~Z. Sheng, H.~Xiong, and L.~Akoglu,
  ``A comprehensive survey on graph anomaly detection with deep learning,''
  \emph{IEEE Transactions on Knowledge and Data Engineering}, 2021.

\bibitem{liang2020feedergan}
M.~Liang, Y.~Meng, J.~Wang, D.~L. Lubkeman, and N.~Lu, ``Feedergan: Synthetic
  feeder generation via deep graph adversarial nets,'' \emph{IEEE Transactions
  on Smart Grid}, vol.~12, no.~2, pp. 1163--1173, 2020.

\bibitem{do2009forecasting1}
M.~B. Do~Coutto~Filho and J.~C.~S. de~Souza, ``Forecasting-aided state
  estimation---part i: Panorama,'' \emph{IEEE Transactions on Power Systems},
  vol.~24, no.~4, pp. 1667--1677, 2009.

\bibitem{do2009forecasting2}
M.~B. Do~Coutto~Filho, J.~C.~S. de~Souza, and R.~S. Freund, ``Forecasting-aided
  state estimation---part ii: Implementation,'' \emph{IEEE Transactions on
  Power Systems}, vol.~24, no.~4, pp. 1678--1685, 2009.

\bibitem{zhang2019power}
L.~Zhang, G.~Wang, and G.~B. Giannakis, ``Power system state forecasting via
  deep recurrent neural networks,'' in \emph{ICASSP 2019-2019 IEEE
  International Conference on Acoustics, Speech and Signal Processing
  (ICASSP)}.\hskip 1em plus 0.5em minus 0.4em\relax IEEE, 2019, pp. 8092--8096.

\bibitem{hossain2021state}
M.~J. Hossain and M.~Rahnamay-Naeini, ``State estimation in smart grids using
  temporal graph convolution networks,'' in \emph{2021 North American Power
  Symposium (NAPS)}.\hskip 1em plus 0.5em minus 0.4em\relax IEEE, 2021, pp.
  01--05.

\bibitem{boyaci2021joint}
O.~Boyaci, M.~R. Narimani, K.~R. Davis, M.~Ismail, T.~J. Overbye, and
  E.~Serpedin, ``Joint detection and localization of stealth false data
  injection attacks in smart grids using graph neural networks,'' \emph{IEEE
  Transactions on Smart Grid}, vol.~13, no.~1, pp. 807--819, 2021.

\bibitem{wang2020locational}
S.~Wang, S.~Bi, and Y.-J.~A. Zhang, ``Locational detection of the false data
  injection attack in a smart grid: A multilabel classification approach,''
  \emph{IEEE Internet of Things Journal}, vol.~7, no.~9, pp. 8218--8227, 2020.

\bibitem{wang2021kfrnn}
Y.~Wang, Z.~Zhang, J.~Ma, and Q.~Jin, ``Kfrnn: an effective false data
  injection attack detection in smart grid based on kalman filter and recurrent
  neural network,'' \emph{IEEE Internet of Things Journal}, vol.~9, no.~9, pp.
  6893--6904, 2021.

\bibitem{horn2012matrix}
R.~A. Horn and C.~R. Johnson, \emph{Matrix analysis}.\hskip 1em plus 0.5em
  minus 0.4em\relax Cambridge university press, 2012.

\bibitem{ramakrishna2019modeling}
R.~Ramakrishna and A.~Scaglione, ``On modeling voltage phasor measurements as
  graph signals,'' in \emph{2019 IEEE Data Science Workshop (DSW)}.\hskip 1em
  plus 0.5em minus 0.4em\relax IEEE, 2019, pp. 275--279.

\bibitem{anis2016efficient}
A.~Anis, A.~Gadde, and A.~Ortega, ``Efficient sampling set selection for
  bandlimited graph signals using graph spectral proxies,'' \emph{IEEE
  Transactions on Signal Processing}, vol.~64, no.~14, pp. 3775--3789, 2016.

\bibitem{zimmerman2010matpower}
R.~D. Zimmerman, C.~E. Murillo-S{\'a}nchez, and R.~J. Thomas, ``Matpower:
  Steady-state operations, planning, and analysis tools for power systems
  research and education,'' \emph{IEEE Transactions on power systems}, vol.~26,
  no.~1, pp. 12--19, 2010.

\bibitem{kipf2017semi}
T.~N. Kipf and M.~Welling, ``Semi-supervised classification with graph
  convolutional networks,'' in \emph{International Conference on Learning
  Representations (ICLR)}, 2017.

\end{thebibliography}
\end{document}